\documentclass[letterpaper]{article} 
\usepackage[draft]{aaai25}  
\usepackage{times}  
\usepackage{helvet}  
\usepackage{courier}  
\usepackage[hyphens]{url}  
\usepackage{graphicx} 
\usepackage{natbib}  
\usepackage{caption} 
\usepackage{algorithm}
\usepackage{algorithmic}

\usepackage{newfloat}
\usepackage{listings}
\DeclareCaptionStyle{ruled}{labelfont=normalfont,labelsep=colon,strut=off} 
\floatstyle{ruled}
\newfloat{listing}{tb}{lst}{}
\floatname{listing}{Listing}
\usepackage{amsmath}
\usepackage{amssymb}
\usepackage{booktabs}

\title{AMG: Avatar Motion Guided Video Generation}
\author{
    Zhangsihao Yang\textsuperscript{\rm1}, Mengyi Shan\textsuperscript{\rm2} \\
    Mohammad Farazi\textsuperscript{\rm1}, Wenhui Zhu\textsuperscript{\rm1}, Yanxi Chen\textsuperscript{\rm1}, Xuanzhao Dong\textsuperscript{\rm1}, Yalin Wang\textsuperscript{\rm1}
}
\affiliations{
    \textsuperscript{\rm 1}Arizona State University\\
    \textsuperscript{\rm 2}University of Washington
}

\usepackage{bibentry}

\begin{document}

\maketitle

\begin{abstract}

Human video generation task has gained significant attention with the advancement of deep generative models. Generating realistic videos with human movements is challenging in nature, due to the intricacies of human body topology and sensitivity to visual artifacts.
The extensively studied 2D media generation methods take advantage of massive human media datasets, but struggle with 3D-aware control; whereas 3D avatar-based approaches, while offering more freedom in control, lack photorealism and cannot be harmonized seamlessly with background scene. We propose \textit{AMG}, a method that combines the 2D photorealism and 3D controllability by conditioning video diffusion models on controlled rendering of 3D avatars. 
We additionally introduce a novel data processing pipeline that reconstructs and renders human avatar movements from dynamic camera videos. 
AMG is the first method that enables multi-person diffusion video generation with precise control over camera positions, human motions, and background style. 
We also demonstrate through extensive evaluation that it outperforms existing human video generation methods conditioned on pose sequences or driving videos in terms of realism and adaptability.

\end{abstract}

\begin{links}
    \link{Code \& Dataset}{https://github.com/zshyang/amg}
\end{links}

\section{Introduction}

Recent progresses in deep generative models have significantly advanced the progress of creative content generation, especially media including images and videos~\cite{blattmann2023stable}. Among those advancements, generating controllable and photorealistic human videos emerges as a particularly interesting and challenging task, with broad potential application in fields including VR/AR, film and game industries. Compared with general open-domain text-to-video generation studies, human video generation is exceptionally hard due to human bodies' special topological restrictions, and the high sensitivity to artifacts that can easily be perceived in the output.

\begin{figure} [!ht]
\includegraphics[width=\linewidth]{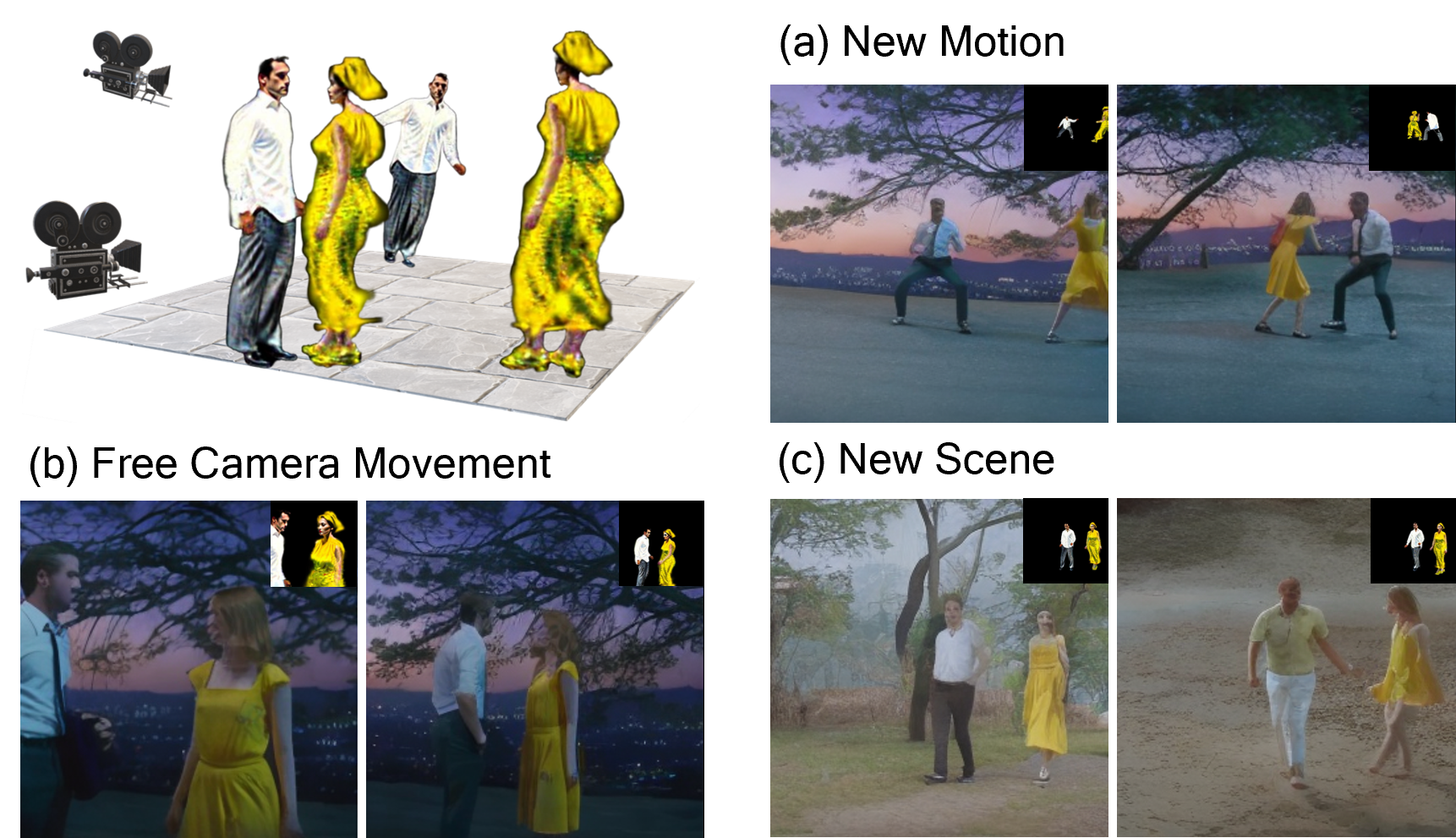}
\centering
\caption{
Our proposed method generates realistic human videos given a single text prompt. We enables diverse controls by explicitly incorporating the rendering of a 3D human avatar as conditional signal while fine-tuning a pre-trained video model. We specially achieves generation with (a) novel motion by generating control motion sequences with various text prompts, (b) free camera viewpoints by simulating camera movements while rendering, and (c) novel scene by describing background in the prompt. 
}
\label{fig:teaser}
\end{figure}

Current human content generation methods mainly develops in two separate streams depending on whether they are 2D or 3D in nature. The 2D methods takes advantage of tremendous scale human media datasets and universal pre-trained generative models. They fine-tune existing image or video models on a specific human dataset, with human pose sequences~\cite{guler2018densepose, SMPL:2015,SMPL-X:2019} as control signals~\cite{xu2023magicanimate, hu2023animateanyone, zhu2024champ}. On the other hand, the 3D methods~\cite{liu2023humangaussian, liao2024tada, kolotouros2023dreamhuman, huang2023humannorm, hong2022avatarclip, zhang2023avatarverse} start by creating animatable human avatars with certain geometry and texture, and later on driving them with user-specified motions. Despite the remarkable progress, neither of these two lines of works satisfies the demands for precise controllability and true photorealism. 2D approaches struggles with full, 3D-aware controllability with dynamic camera positions, detailed 3D appearance information, and group interactive motions that often involves body occlusions. 
Specifically, skeletal signals like OpenPose~\cite{hu2023animateanyone,wang2023disco} struggle with handling occlusion, while semantic signals like DensePose~\cite{xu2023magicanimate,karras2023dreampose} face challenges in preserving identity.
3D approaches often emphasize computing lighting transitions between objects and scenes but lack methods to leverage the massive amount of photorealistic data contents. This limitation results in unnatural rendering or ignorance of the background, particularly in scenarios requiring realistic lighting, which in turn demands high computational resources.

In this paper, we propose \textit{AMG}, a method that merges merits from both 2D and 3D worlds. In particular, we introduce a technique that uses controlled 3D avatar rendering as a condition for video diffusion models.  
However, creating a dataset with paired 3D animatable human avatars and photorealistic videos is not a trivial task, as existing datasets lack such pairs. 
There are two approaches to address this challenge: either reconstruct a 3D human avatar from 2D video datasets or first render an albedo-like video and then refine the avatar with modern rendering techniques to achieve photorealism. We choose the first approach due to the abundance of 2D videos and the maturity of 3D human reconstruction from 2D videos.
Given a human video in training data, we start by building an animatable 3D human avatar~\cite{liu2023humangaussian} for identities occurring in the video with prompts generated from Vision Language Model (VLM)~\cite{liu2023llava} describing their appearance. We then extract 3D human body movements from the same video, use the pose sequences to drive the synthesized avatars, and simulate cameras to render synthetic videos of the human avatars. This synthetic rendering serves as the conditional signal when we fine-tune a text-to-video model our human video datasets. In particular, we employ our collected dataset of \texttt{(video, prompt, avatar rendering)}, add additional condition by concatenating the frame with noise in latent space~\cite{brooks2023instructpix2pix} and fine-tune the diffusion model ModelScopeT2V~\cite{wang2023modelscope} with LoRA~\cite{hu2021lora}. 

At inference time, we take a single text prompt as input and decompose it into human appearance and scene descriptions.
we  synthesize the human avatar~\cite{liu2023humangaussian} based on the appearance description and then use off-the-shelf multi-person motion generation models~\cite{liang2024intergen,shan2024towards} to generate driving signals for the avatars.
The synthetic videos and scene descriptions are then used to condition our video diffusion model.
Our model design enables a high-level of controllability that combines the strengths from both the powerful 2D pre-trained model, and the rich 3D information of human avatars. Figure~\ref{fig:teaser} visualizes various applications of our method. 
Remarkable, our method is the first to achieve multi-person diffusion video generation, allowing for a high degree of controllability over multiple unexplored aspects such as camera positions, human 3D motions, and lighting adjustments to match the described scene.

To summarize, our contributions are as follows: (a) We propose a training paradigm that takes advantage of both 2D pre-trained generative models and 3D animatable avatars, achieving high-level of realism as well as fine-grained controllability for human video generation. 
(b) We introduce a data collection pipeline that extracts 3D motion and camera information from 2D human videos, and render corresponding human avatar as conditions. We also plan to release the processed dataset. 
(c) We propose a video conditional parameter-efficient fine-tuning method for pre-trained text-to-video models.
(d) We compare our method against multiple baselines and verify its effectiveness, robustness, and superiority in terms of both controllability and photorealism, as measured by CLIP score and motion accuracy.

\section{Related Work}

\noindent\textbf{Text-to-Video Generation}

\noindent The text-to-video generation task aims to synthesize plausible, text-aligned, and temporally coherent video sequences given a prompt description. Early techniques including combining RNN and GANs~\cite{yu2022digan, skorokhodov2021styleganv, tulyakov2018mocogan}, and Transformer-based autoregressive models~\cite{yang2024cogvideox, hong2022cogvideo, yu2023magvit}. Recent progress in diffusion models greatly boosts the advancement in video generation~\cite{xing2023dynamicrafter, he2022lvdm}. Early works like Video Diffusion Model~\cite{ho2022video}, ModelScopeT2V~\cite{wang2023modelscope}, and VideoCrafter~\cite{chen2023videocrafter1, chen2024videocrafter2} start with the innovative idea of factorizing space and time to improve efficiency, and thus allow models to build upon existing Text-to-Image model. Video LDM~\cite{blattmann2023videoldm} and EMU video~\cite{girdhar2024emu} learns to augment pre-trained U-Net structure image generation model with temporal layer and 3D convolution, and jointly train with videos and images. Imagen-Video~\cite{ho2022imagenvideohighdefinition} learns a series of cascaded video diffusion models that starts with initial short and low-resolution video, and then extending to long high-resolution ones with spatial and temporal upsampling. More recent work explores the creative idea of replacing the U-Net structure with Transformer architecture~\cite{gupta2023walt, ma2024latte}, inspired by the promising Text-to-Image generative results from DiT~\cite{peebles2022dit}. Our work is built upon ModelScopeT2V~\cite{wang2023modelscope}, and it can be seamlessly transferred to other Text-to-Video models. 

\subsection{Controllable Video Generation}
Controllable video generation tasks leverages guidance from content control signals other than text. Example control modalities include semantic story layout~\cite{gong2023talecrafter, long2024videodrafter}, sketches~\cite{gal2023breathing}, optical flow~\cite{yang2023rerender, hu2023videocontrolnet}, and camera movements~\cite{xu2024camco, yang2024direct, he2024cameractrl}.~\cite{guo2023sparsectrl} combines various control types above to train a universal control video generation model. One line of controllable video generation work studies the motion in a video. DragNUWA~\cite{yin2023dragnuwa} enables high-level control with text, image and user-defined trajectory to control both object movements and camera path. Generative Image Dynamics~\cite{li2024generative} models scene motion from a collection of motion trajectories extracted from real video sequences. DMT~\cite{yatim2024space} tackles the video translation task, using a pre-trained text-to-video diffusion model to transform video scenes according to text prompts while preserving the original motion. AnimateAnything~\cite{dai2023animateanything} animates a reference image by designating an area of motion within a 2D image. ~\cite{wang2024boximator} enables fine-grained motion control with user-input bounding box and trajectories. 

\begin{figure*} [t]
\includegraphics[width=\linewidth]{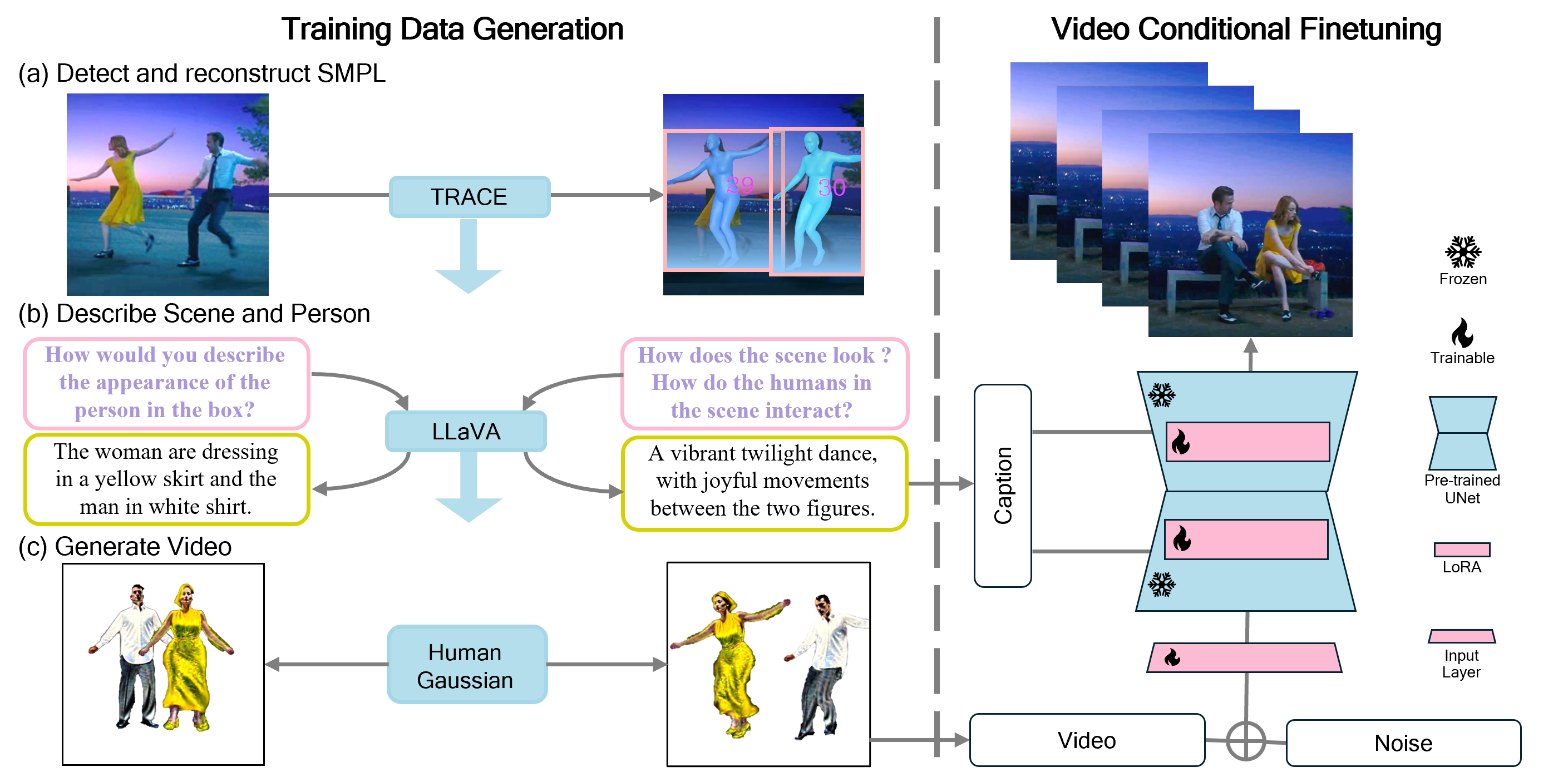}
\centering
\caption{
Our method consists of two stages: training data generation and video-conditional finetuning. 
In the left column, we visualize key steps in our data generation pipeline. We begin by (a) detecting and reconstructing SMPL using TRACE; then (b) using LLaVA to generate textual descriptions that capture both the subjects' appearance and their interaction with the environment; and finally
(c) rendering an avatar video using HumanGaussian, based on motion and camera from (a) and appearance from (b). 
In the right column, we illustrate how the synthetic human avatar video is used to condition the fine-tuning process by leveraging the input video condition and LoRA.
}
\label{fig:data}
\end{figure*}

\noindent\textbf{Human Video Generation}

\noindent As the study of general object and scene motion grows, a specific subset of controllable human motion-conditioned video generation also evolves simultaneously. Human video generation is in nature hard, especially due to specific body topology and audiences' high level of awareness to even the smallest artifact. Existing method utilizes motion guidance to improve video faithfulness, signals including OpenPose skeletons~\cite{hu2023animateanyone, wang2023disco}, DensePose~\cite{xu2023magicanimate, karras2023dreampose}, SMPL~\cite{zhu2024champ}, or simply a driving video~\cite{yatim2024space}. It still remains challenging to generate videos that allow for detailed 3D control of human body movements and appearance. 
Our method instead takes explicit 3D human avatar rendering as the control signal for video diffusion models, thus achieving finer level of controllability, and also enables multi-person interactive video generation.

\noindent\textbf{Human Avatar Generation from Text}

\noindent Realistic 3D human generation from text prompts is an emerging and challenging task. 
Current research mainly focuses on two aspects to address this challenge: underlying 3D representations and supervisions to optimize them. 
The 3D representations can be based on either 2D manifolds (such as SMPL~\cite{SMPL:2015} and SMPL-X~\cite{SMPL-X:2019}), or 3D volumetric fields (such as NeuS~\cite{wang2021neus}, imGHUM~\cite{alldieck2021imGHUM}, and 3D Gaussian Splatting~\cite{kerbl20233Dgaussians}). 
Supervisions are categorized into diffusion model-guided approaches (also known as Score Distillation Sampling~\cite{poole2022dreamfusion}), which are trained on various conditions (DensePose~\cite{guler2018densepose}, normal maps, and depth maps), and non-diffusion-guided approaches (CLIP~\cite{radford2021learning}). 
The combination of these two aspects leads to the development of multiple methods, including AvatarCLIP~\cite{hong2022avatarclip}, DreamHuman~\cite{kolotouros2023dreamhuman}, AvatarVerse~\cite{zhang2023avatarverse}, TADA~\cite{liao2024tada}, HumanNorm~\cite{huang2023humannorm}, and HumanGaussian~\cite{liu2023humangaussian}.
Despite the rapid growth in 3D human quality, none of the existing approaches achieve photo-realistic quality, especially when rendering with various camera poses, user-given motion sequences and backgrounds.
Our method enables control over video generation using human avatars while also enhancing avatar rendering quality.

\section{Method}

\noindent\textbf{Preliminaries}

\noindent\underline{Text-to-Video Diffusion Models} are designed to transform textual input into a video data distribution using a reverse diffusion process~\cite{ho2020denoising,sohl2015deep}.
These models typically operate in a latent space to efficiently manage the complexity of video data~\cite{rombach2022high}.
During pre-training, a video sample $x$ is encoded by a pre-trained encoder $\mathcal{E}$~\cite{esser2021taming} to obtain its latent representation $z \in \mathbb{R}^{f \times h \times w \times 4}$.
In the forward diffusion process, random noise $\epsilon$ is added to $z$ according to a pre-defined noise schedule $\{\beta_t\}_{t=1}^T$. 
This process is represented as $z_t = \sqrt{\bar{\alpha}_t}z + \sqrt{1 - \bar{\alpha}_t}\epsilon$, where $\epsilon \sim \mathcal{N}(0,1)$ represents Gaussian noise with the same dimensions as $z$, $\bar{\alpha}_t = \prod_{s=1}^t \alpha_s$, and $\alpha_t = 1 - \beta_t$. 
A UNet~\cite{ronneberger2015u,cciccek20163d} model, denoted as $\epsilon_\theta$, is employed to denoise $z_t$, enabling video generation through the reverse diffusion process, conditioned on the video caption $c$. 
The optimization is guided by the following reweighted variational bound~\cite{ho2020denoising}:
\begin{equation}
\label{eq:vb}
\mathcal{L}(\theta) = \mathbb{E}_{z, c, \epsilon, t} \left[ \left\| \epsilon - \epsilon_\theta \left(\sqrt{\bar{\alpha}_t}z + \sqrt{1 - \bar{\alpha}_t}\epsilon, c, t\right) \right\|_2^2 \right]
\end{equation}
For video generation during inference, the DDIM sampling method~\cite{song2020denoising} is utilized to produce video outputs.

\noindent\underline{Low-Rank Adaptation (LoRA)} finetuning~\cite{hu2021lora} is an approach that reduces memory requirements by introducing a small set of trainable parameters, commonly referred to as adapters, while keeping the full model frozen. 
During stochastic gradient descent, the gradients are propagated through the fixed pretrained model weights to update only the adapters. 
LoRA achieves this by augmenting a linear projection with an additional factorized projection. 
Specifically, given a projection $\mathbf{XW} = \mathbf{Y}$ where $\mathbf{X} \in \mathbb{R}^{b \times h}$ and $\mathbf{W} \in \mathbb{R}^{h \times o}$, LoRA modifies the projection as follows:
\begin{equation}
\mathbf{Y} = \mathbf{XW} + s\mathbf{XL}_1\mathbf{L}_2,
\end{equation}
where $\mathbf{L}_1 \in \mathbb{R}^{h \times r}$, $\mathbf{L}_2 \in \mathbb{R}^{r \times o}$, and $s$ is a scalar.

\noindent\textbf{Creating an Avatar-Motion Video Dataset}

\noindent We aim to create a dataset $\{(V, y_s, V_a)\}$, where $V$ represents a real human video clip, $y_s$ is a textual description of the video scene, and $V_a$ is a video composed of frames $I_a$ that are rendered using synthetic, articulated avatars. 
These avatars are driven by the motion detected from $V$.

\noindent\underline{Extract SMPL poses.}
We begin by utilizing TRACE~\cite{sun2023trace} to locate, track and extract the SMPL~\cite{SMPL:2015} parameters of human bodies in $V$. 
TRACE processes $V$ and generates the SMPL body pose parameters $\theta \in \mathbb{R}^{24 \times 6}$ as well as global camera parameters.

\noindent\underline{Generate appearance prompts.}
To generate the caption \( y_a \) that describes the appearance of the human in \( V \) for creating 3D human avatars, we begin by manually selecting a key frame from \( V \) that clearly exhibits the full body appearance of the characters. 
We compute a 2D bounding box based on the bounding box of the generated mesh $M$, which is derived from the SMPL parameters $\theta$. 
This bounding box is then overlaid onto the original image \( I \), producing the image \( I_b \). 
Next, we ask LLaVA~\cite{liu2023llava} to describe the appearance of the character by posing the question \( Q_a \): "How would you describe the appearance of the person in the bounding box?".
The output from LLaVA $y_a$, is used as the descriptive prompt. Refer to the left most column of Figure~\ref{fig:data} for a visual illustration of this pipeline.

\noindent\underline{Animate 3DGS.}
We use $y_a$ as the prompt for HumanGaussian~\cite{liu2023humangaussian}, a method that leverages a pretrained 2D model to extract information for training a textured A-posed articulated 3DGS $g_{\phi}$. 
We animate the synthesized HumanGaussian avatar $g_{\phi}$ with the SMPL parameters $\theta$ extracted by TRACE to generate a sequence of 3D avatar movements that takes both appearance and motion into consideration. Then we render this 3D sequence into 2D videos by simulating cameras following the parameters estimated by TRACE along with the SMPL poses. We denote the rendered image with black background $I_a = g_{\phi}(\theta)$.

\noindent\underline{Generate scene prompts.}  
The final step is to generate the caption $y_s$ that describes the scene in $V$. 
We use the middle frame of $V$ as input to LLaVA, posing the question $Q_s$: "How does the scene look? How do the humans in the scene interact with each other and the environment? What is the atmosphere of the image?".
The output $y_s$ from LLaVA serves as the scene description.

In summary, our data processing module takes a real video $V$ with multiple interactive people as input, and outputs a synthetic video $V_a$, where the realistic human motions are transformed into avatars rendering, along with the scene description $y_s$. 
The off-the-shelf modules we used, including TRACE (video-to-motion), LLaVA (vision-language QA), and HumanGaussian (text-to-3D avatar) can be replaced by any advanced pre-trained methods with similar functionality, demonstrating the flexibility of the approach.
More data processing details are in the Appendix.

\begin{figure} [t]
\includegraphics[width=\linewidth]{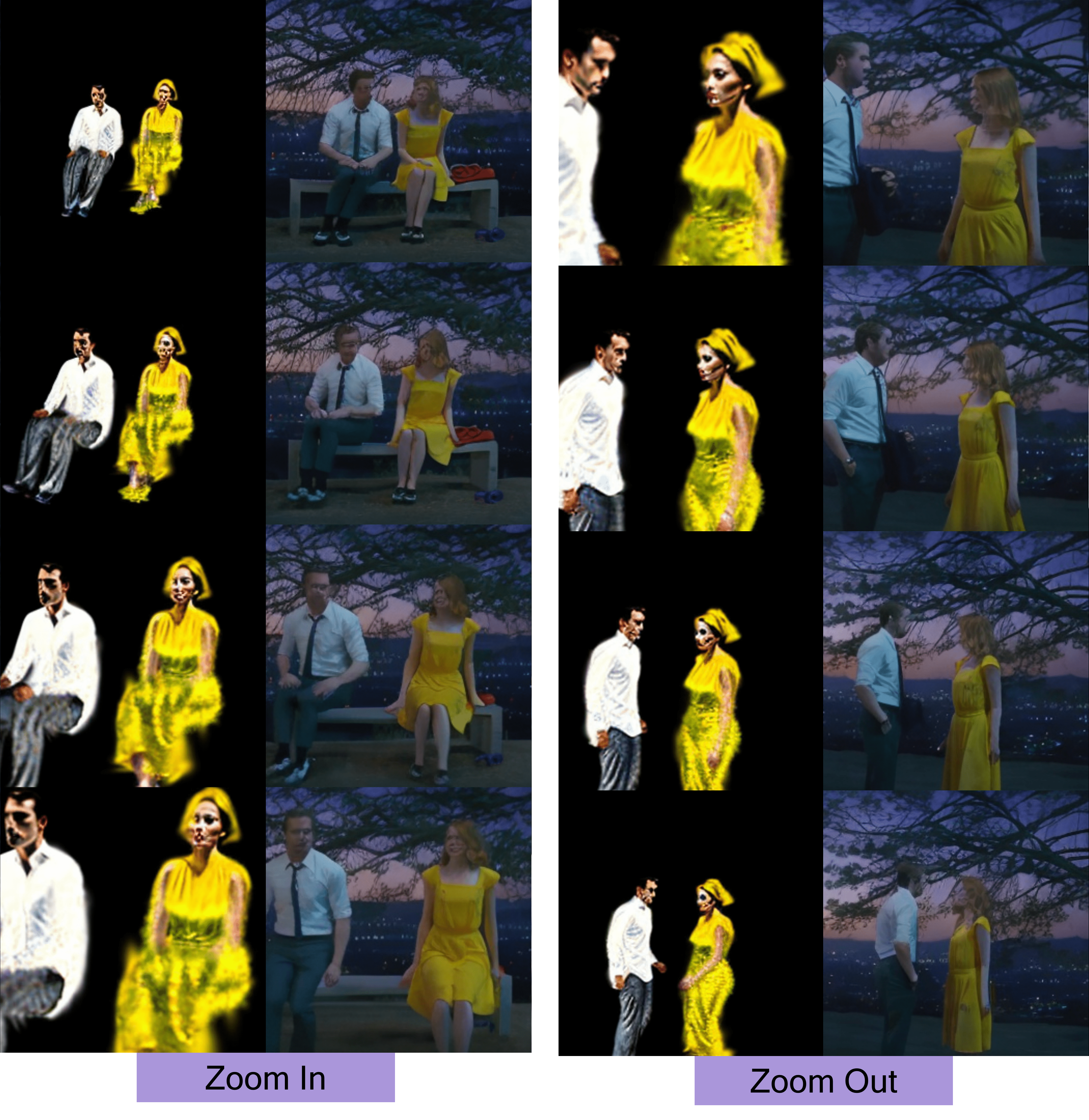}
\centering
\caption{
Video with explicit camera movement control. 
The left column shows a zoom-in sequence, and the right column shows a zoom-out sequence. 
Each column pairs the input rendered avatar video on the left with the generated video from our method on the right.
}
\label{fig:camera}
\end{figure}
\begin{figure*} [!ht]
\includegraphics[width=\linewidth]{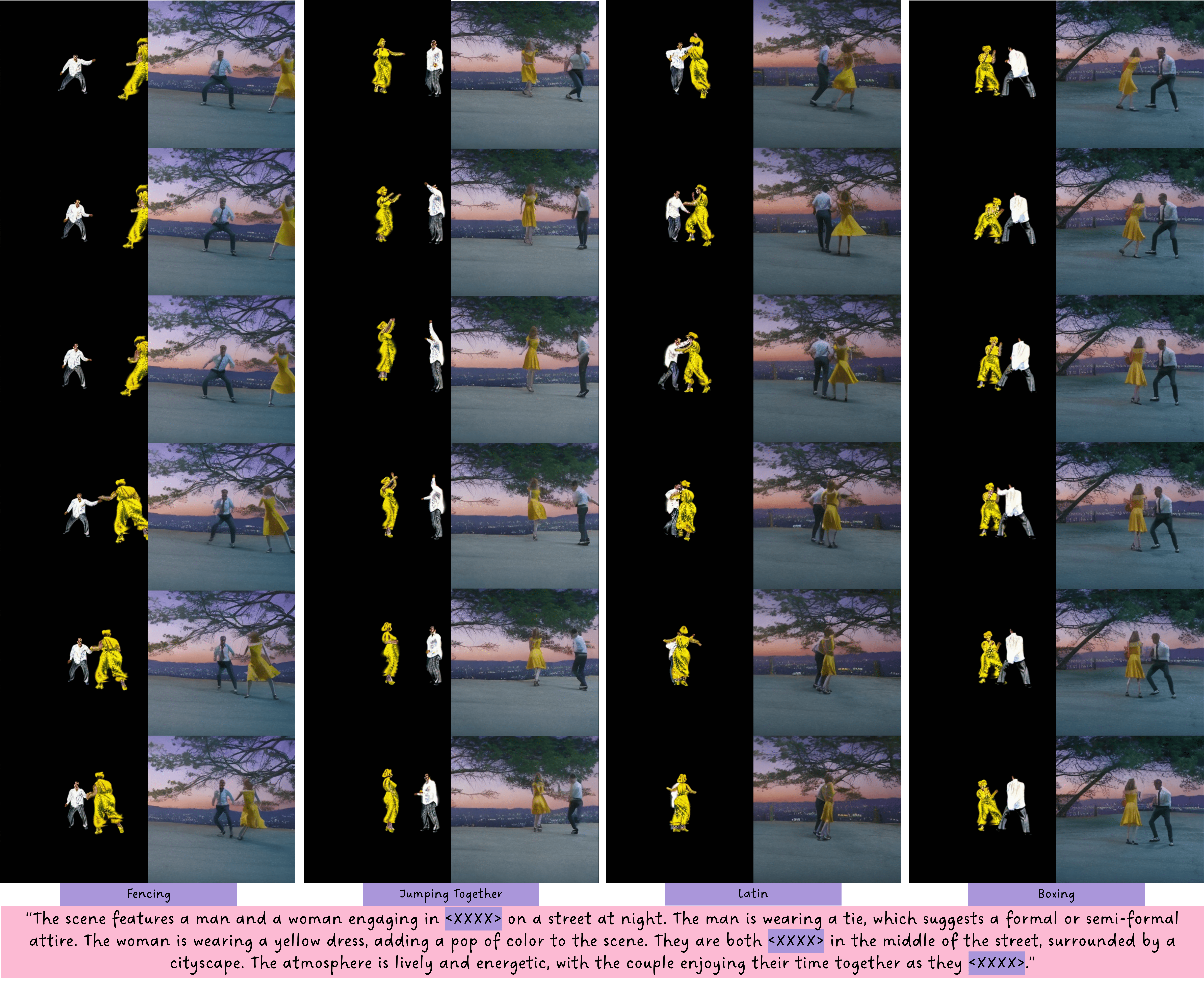}
\centering
\caption{
Video with user-defined human motion control. Given an action prompt, we start with animating the pre-generated human avatar by motions generated from the text. We render the motions for a specific camera angle (left columns in each pair), and feed that as a condition to our video model to generate photorealistic human videos. Our model is able to generate videos of various novel activities that are out of distribution of the original training data.
}
\label{fig:motion}
\end{figure*}

\noindent\textbf{Model Video Conditional Tuning}

\noindent We train a text and video-conditioned diffusion model, which leverages LoRA to generate video from $V_a$ and $y_s$. 
Our model is built upon ModelScopeT2V~\cite{wang2023modelscope}, a large-scale text-to-video latent diffusion model.

Compared to the text-to-video objective in Eq.~\ref{eq:vb}, we introduce an additional condition, $z_a = \mathcal{E}(V_a)$, into the model. 
The corresponding latent diffusion objective is as follows:
\begin{equation}
\label{eq:vbc}
\mathcal{L}(\theta) = \mathbb{E}_{z, y_s, z_a, \epsilon, t} \left[ \left\| \epsilon - \epsilon_\theta \left(z_t, y_s, z_a, t\right) \right\|_2^2 \right]
\end{equation}

Previous works including PITI~\cite{wang2022pretraining} and InstructPix2Pix~\cite{brooks2023instructpix2pix} demonstrate that fine-tuning large image diffusion models often outperforms training a model from scratch for image translation tasks, particularly when paired training data is limited. 
Consequently, we initialize the weights of our model using a pretrained ModelScopeT2V~\cite{wang2023modelscope} checkpoint, leveraging its extensive text-to-video generation capabilities for the non-LoRA components, which are subsequently frozen. For the LoRA components, $\mathbf{L_1}$ is initialized randomly, while $\mathbf{L_2}$ is initialized with zeros.

To enable video conditioning, we double the size of the input channels to the first convolutional layer. For each frame, we concatenate the noise latent $z_t$ and the additional condition frame $\mathcal{E}(V_a)$ on the channel dimension, and pass it through the enlarged first layer. For text-conditioning, we reuse the original cross-attention conditioning mechanism, which injects the CLIP embedding of caption into the module through cross-attention transformers. 

All accessible weights of the diffusion model are initialized from the pretrained ModelScopeT2V checkpoints, whereas the weights corresponding to the newly added input channels are initialized to zero. Specifically, we freeze the CLIP embedder for text condition.

\noindent\textbf{Inference}

\noindent At inference time, we first use a textual description to synthesize two-person interactive motions in SMPL~\cite{SMPL:2015} format with InterGen~\cite{liang2024intergen}. 
We then generate HumanGaussian~\cite{liu2023humangaussian} avatars following the appearance prompts, drive them with the synthesized motions, and control the camera to render frames of detailed movements. 
Finally, we condition our text-to-video model with the rendered motion frames to generate photorealistic human videos that follow both the appearance and movements of the avatar.

\section{Results}

\noindent\textbf{Dataset Preparation and Implementation Details}

\noindent We collect our film video dataset which contains 6,771 frames, recorded at 24 frames per second (FPS). We filter out clips where the subject is occasionally lost or absent due to TRACE's tracking issues, and reduce the dataset to 5,788 valid clips, with a downsampled FPS of 8. 
We use the first 5,500 clips for training, keeping and the rest for testing the performance of our method and baseline models.
Further details can be found in the Appendix.

For our experiments, we set the LoRA rank to 4 and kept both the spatial and temporal gradient weights at 1. We adopt the publicly available text-to-video diffusion model ModelScopeT2V as the base model. ModelScopeT2V is pre-trained on WebVid10M~\cite{Bain21} with 1000 DDIM steps and is capable of generating videos at a resolution of $16 \times 256 \times 256$. We performed a 20-step DDIM inference, incorporating classifier-free guidance~\cite{ho2022classifier} by default. 
Experiments were conducted on 8 NVIDIA A100 GPUs, with a batch size of 8 and a learning rate of $1 \times 10^{-6}$. 
To balance cost and performance, we fine-tuned AMG with default parameters for 60k steps, unless otherwise specified.
For the reasoning behind using 60k steps, the Appendix provides a detailed explanation.

\noindent\textbf{Qualitative Evaluation}

\noindent\underline{Motion Modification.}
Our model can generalize to novel, out-of-domain (OOD) motions that never appear in the training data, and generate corresponding videos. 

In Fig.~\ref{fig:motion}, we present examples of various OOD actions, including boxing, fencing, Latin dance, and jumping. Thanks to the power of the off-the-shelf text-to-motion model, we are able to generate interactive human motions used to drive the avatar, and condition the video model on the rendered videos. 
First, the model accurately captures the motion locations.
Additionally, the model successfully follows the input conditions, reflecting detailed motions that are often difficult to describe in text prompts but can be replicated by graphic artists.
Notably, the model captures subtle details effectively. 
For instance, in the fencing example, the squatting motion is accurately reflected in the generated video.
In the Latin dance example, the model demonstrates a strong understanding of complex depth exchanges, producing high-quality video outputs. 
We attribute this capability to the presence of dance-related data in the training data.
For additional analysis, please refer to the Appendix.

\noindent\underline{Camera Movement.}
In addition to human motion, camera movement plays a critical role in the dynamics of 3D scenes. While it is often ignored by general 2D generative model without fine-tuning on annotated datasets, our proposed pipeline is able to fully control the camera trajectory when rendering 3D human avatars. 
This gives us the freedom to manage camera movement in 3D space. 
In Fig.~\ref{fig:camera}, we present video generated under two camera movement scenarios: zooming in and zooming out. 
The results indicate that the generated video can faithfully follow the user-specified camera movements while simultaneously capturing the avatars' motions. 
Additionally, non-human objects in the generated video, such as the stone bench, also accurately respond to the camera movements, demonstrating that camera dynamics priors are preserved from the pre-trained model. 
For more examples, please refer to the Appendix.

\begin{figure} [t]
\includegraphics[width=\linewidth]{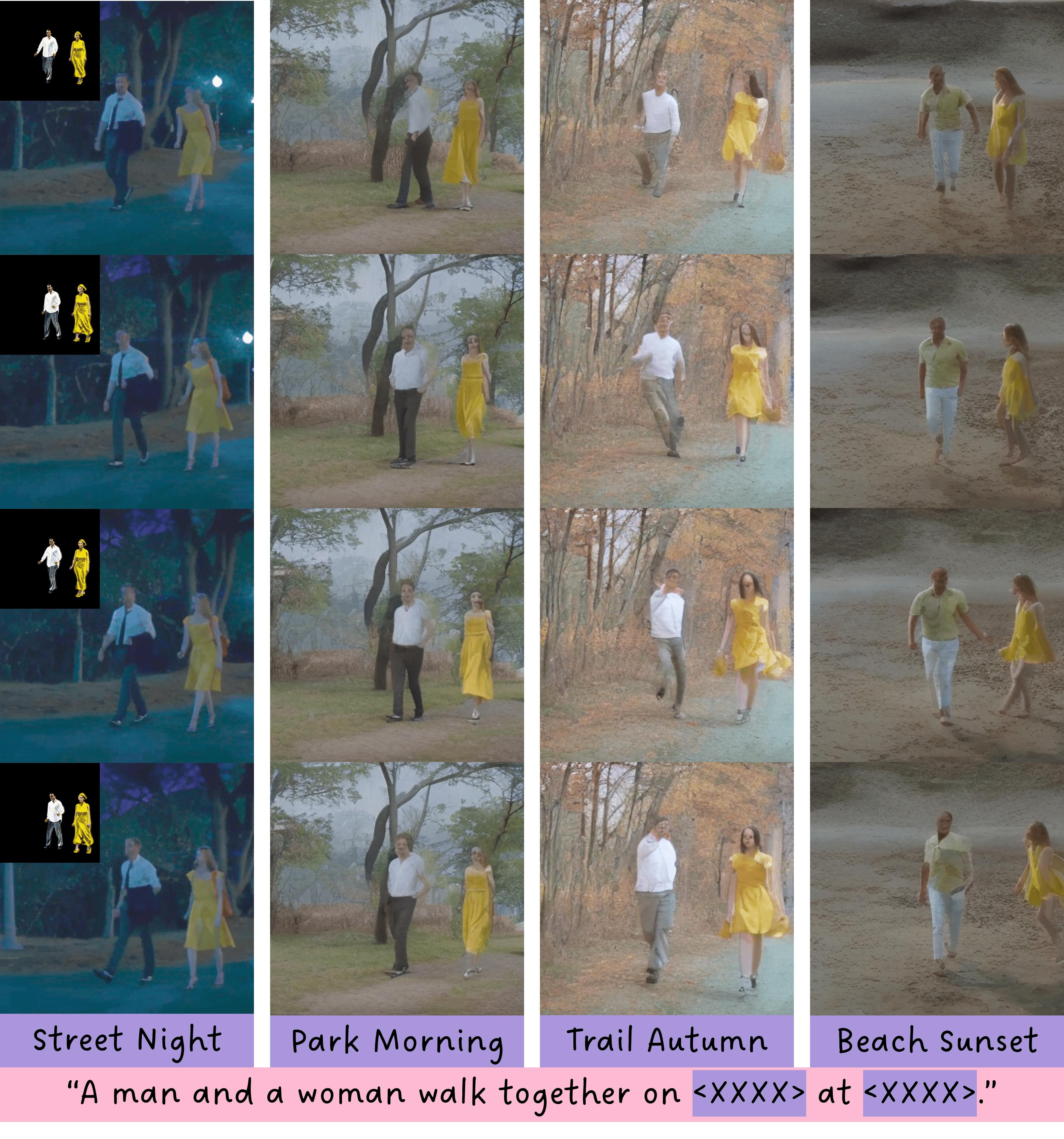}
\centering
\caption{
Video with background changes. First column's left upper corner presents the same character avatar rendering, and the rest shows the result video with different prompts describing the scene. 
}
\label{fig:background}
\end{figure}
\begin{figure*} [t]
\includegraphics[width=\linewidth]{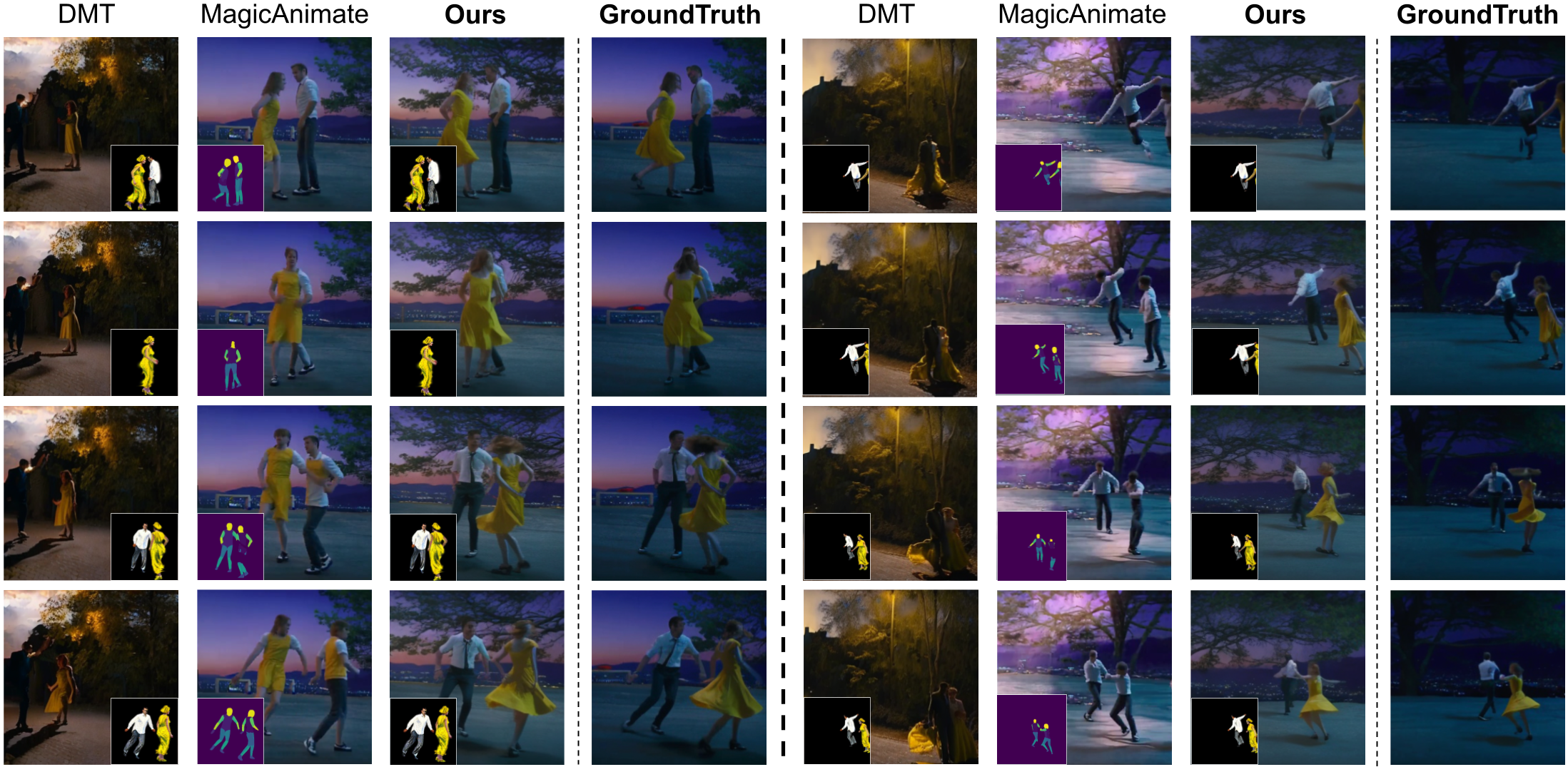}
\centering
\caption{Visual comparison with baselines. Note that DMT is unable to generate background or character appearance following the prompt. MagicAnimate takes the first ground-truth frame as reference, but still messes up the identities when there are more than one subjects with interactive motions. 
}
\label{fig:baseline}
\end{figure*}

\noindent\underline{Background Modification.}
After the LoRA-based fine-tuning stage, our model still preserves the general knowledge from the pre-trained model and can achieves effects of background modification following changes in prompts.

As shown in Fig.~\ref{fig:background}, we keep the human avatars while changing the scene prompt for video generation. Our results indicate that the model successfully preserves the identity features, such as the appearance and gender of the input, while effectively generalizing to different backgrounds. The generated video also maintains the consistent human locations and movements with the input avatar rendering.  

Notably, when the background is switched to a beach setting, the model adapts the clothing, changing the shoes and shorts accordingly. It is additionally able to seamlessly modify the lighting and color tone for generated human to harmonize with the background scene. Such realism validates that our model not only takes advantage of rich 3D-aware condition, but enjoys the strong 2D prior from the pre-trained model retained by LoRA. 

\noindent\textbf{Baselines}

\noindent We compare our method with two representative categories of current works: those directly using video as a guidance for motion signal, and those using human motion sequences as intermediate control signals. For the former category, we compare with a representative work DMT~\cite{yatim2024space} , a training-free model that takes in a video and a textual prompt to achieve the effect of text-driven motion transfer. We use our rendered HumanGaussian outputs $V_a$ as the video input to control motion, and the textual prompt to control the video ``style''. 
We follow the steps in DMT, including DDIM inversion and SMM optimization sampling.
For the latter category, we compare with a pioneering work on human video generation controlled by a reference frame and a pose sequence, namely MagicAnimate~\cite{xu2023magicanimate}. We extract pose sequences from the ground-truth video, and use the first real video frame as reference frames. Note that this evaluation setting is strictly in favor of those baselines, because our method doesn't reply on a ground-truth reference frame nor groundtruth pose sequences. Everything is generated on the fly instead.

The results in Fig.~\ref{fig:baseline} suggest that \underline{DMT} heavily relies on the provided prompt. It keeps the major semantic layout, but cannot generate high-quality video with background aligned with the scene prompt. 
DMT also struggles with the precise positioning of avatars.
In the left column, the scale of the generated human figures is not proportional to the input scale. 
On the right, although the scale is preserved, the relative positions of the white t-shirt and yellow skirt are switched. This suggests that DMT does not effectively utilize input information such as color and avatar layout. In contrast, our method leverages pre-trained text-to-video models, maximizing their potential.
\underline{MagicAnimate} is able to generate reasonable human videos, but cannot identify the two different identities especially when there is some occlusion in the middle. 
As seen in rows 2, 3, and 4, it switches the identities after the man’s body is fully occluded. 
This randomness arises because the method, relying only on semantic signals, fails to maintain consistent identities in overlapping frames. 
Our method, however, maintains identity consistency by directly rendering the corresponding avatars. 

The remaining results are in the Appendix.

\noindent\textbf{Quantitative Evaluation}

\begin{table}[h]
\centering
\begin{tabular}{lcc}
\toprule
\textbf{Method} & \textbf{CLIP Score} $\uparrow$ & \textbf{Motion Score} $\uparrow$   \\ 
\midrule
\textit{Ground Truth}    & 34.43 & 99.88 \\
\midrule
DMT             & 32.51 & 56.31 \\
MagicAnimate    & 33.23 & 26.53 \\
\hline
\textbf{Ours}            & \textbf{33.59}   & \textbf{69.88} \\
\bottomrule
\end{tabular}
\caption{
Quantitative evaluation. Our method achieves the best performance among the methods.
}
\label{tab:baseline}
\end{table}

\noindent In Tab.~\ref{tab:baseline}, we evaluate our method using the CLIP text similarity score and the motion fidelity score, comparing it with other methods. The CLIP score is calculated by measuring the similarity between each frame and the corresponding caption of the clip, with the final score being the average across all frames. The motion fidelity score is determined by evaluating the similarity of tracklets between two videos.
The formal definitions of the two metrics are in the Appendix.

For our evaluation, we select five clips from the withheld test data, each consisting of 16 frames and an FPS of 8, from a total of 288 test clips. Among all the methods, our approach achieves the best overall performance. 

MagicAnimate incorporates the background into the generation process, which contributes to a relatively higher similarity score. In contrast, DMT, which only takes rendered avatars on a black background, is not able to generate the exact scene but produces more accurate motion, resulting in a higher motion fidelity and a lower CLIP similarity score.
Our method effectively generates backgrounds that are aligned with the text, even if they are not identical to the ground truth, and it produces motion that closely matches the ground truth.

\section{Conclusion}
In this work, we propose a novel method to generate human videos following specific appearance and movements. We fine-tune a pre-trained text-to-video diffusion model by conditioning on human avatar movement frames. Our results show superior performance on two-person human video generation than baseline methods. We also demonstrate diverse applications in video generation, generating various backgrounds, novel motions, and free camera poses. 

\bibliography{main}

\begin{thebibliography}{70}
\providecommand{\natexlab}[1]{#1}

\bibitem[{Alldieck, Xu, and Sminchisescu(2021)}]{alldieck2021imGHUM}
Alldieck, T.; Xu, H.; and Sminchisescu, C. 2021.
\newblock imGHUM: Implicit Generative Models of 3D Human Shape and Articulated Pose.
\newblock \emph{2021 IEEE/CVF International Conference on Computer Vision (ICCV)}, 5441--5450.

\bibitem[{Bain et~al.(2021)Bain, Nagrani, Varol, and Zisserman}]{Bain21}
Bain, M.; Nagrani, A.; Varol, G.; and Zisserman, A. 2021.
\newblock Frozen in Time: A Joint Video and Image Encoder for End-to-End Retrieval.
\newblock In \emph{IEEE International Conference on Computer Vision}.

\bibitem[{Blattmann et~al.(2023{\natexlab{a}})Blattmann, Dockhorn, Kulal, Mendelevitch, Kilian, Lorenz, Levi, English, Voleti, Letts et~al.}]{blattmann2023stable}
Blattmann, A.; Dockhorn, T.; Kulal, S.; Mendelevitch, D.; Kilian, M.; Lorenz, D.; Levi, Y.; English, Z.; Voleti, V.; Letts, A.; et~al. 2023{\natexlab{a}}.
\newblock Stable video diffusion: Scaling latent video diffusion models to large datasets.
\newblock \emph{arXiv preprint arXiv:2311.15127}.

\bibitem[{Blattmann et~al.(2023{\natexlab{b}})Blattmann, Rombach, Ling, Dockhorn, Kim, Fidler, and Kreis}]{blattmann2023videoldm}
Blattmann, A.; Rombach, R.; Ling, H.; Dockhorn, T.; Kim, S.~W.; Fidler, S.; and Kreis, K. 2023{\natexlab{b}}.
\newblock Align your Latents: High-Resolution Video Synthesis with Latent Diffusion Models.
\newblock In \emph{IEEE Conference on Computer Vision and Pattern Recognition ({CVPR})}.

\bibitem[{Brooks, Holynski, and Efros(2023)}]{brooks2023instructpix2pix}
Brooks, T.; Holynski, A.; and Efros, A.~A. 2023.
\newblock Instructpix2pix: Learning to follow image editing instructions.
\newblock In \emph{Proceedings of the IEEE/CVF Conference on Computer Vision and Pattern Recognition}, 18392--18402.

\bibitem[{Chen et~al.(2023)Chen, Xia, He, Zhang, Cun, Yang, Xing, Liu, Chen, Wang, Weng, and Shan}]{chen2023videocrafter1}
Chen, H.; Xia, M.; He, Y.; Zhang, Y.; Cun, X.; Yang, S.; Xing, J.; Liu, Y.; Chen, Q.; Wang, X.; Weng, C.; and Shan, Y. 2023.
\newblock VideoCrafter1: Open Diffusion Models for High-Quality Video Generation.
\newblock arXiv:2310.19512.

\bibitem[{Chen et~al.(2024)Chen, Zhang, Cun, Xia, Wang, Weng, and Shan}]{chen2024videocrafter2}
Chen, H.; Zhang, Y.; Cun, X.; Xia, M.; Wang, X.; Weng, C.; and Shan, Y. 2024.
\newblock VideoCrafter2: Overcoming Data Limitations for High-Quality Video Diffusion Models.
\newblock arXiv:2401.09047.

\bibitem[{{\c{C}}i{\c{c}}ek et~al.(2016){\c{C}}i{\c{c}}ek, Abdulkadir, Lienkamp, Brox, and Ronneberger}]{cciccek20163d}
{\c{C}}i{\c{c}}ek, {\"O}.; Abdulkadir, A.; Lienkamp, S.~S.; Brox, T.; and Ronneberger, O. 2016.
\newblock 3D U-Net: learning dense volumetric segmentation from sparse annotation.
\newblock In \emph{Medical Image Computing and Computer-Assisted Intervention--MICCAI 2016: 19th International Conference, Athens, Greece, October 17-21, 2016, Proceedings, Part II 19}, 424--432. Springer.

\bibitem[{Dai et~al.(2023)Dai, Zhang, Yao, Qiu, Zhu, Qin, and Wang}]{dai2023animateanything}
Dai, Z.; Zhang, Z.; Yao, Y.; Qiu, B.; Zhu, S.; Qin, L.; and Wang, W. 2023.
\newblock AnimateAnything: Fine-Grained Open Domain Image Animation with Motion Guidance.
\newblock \emph{arXiv e-prints}, arXiv--2311.

\bibitem[{Esser, Rombach, and Ommer(2021)}]{esser2021taming}
Esser, P.; Rombach, R.; and Ommer, B. 2021.
\newblock Taming transformers for high-resolution image synthesis.
\newblock In \emph{Proceedings of the IEEE/CVF conference on computer vision and pattern recognition}, 12873--12883.

\bibitem[{Gal et~al.(2023)Gal, Vinker, Alaluf, Bermano, Cohen-Or, Shamir, and Chechik}]{gal2023breathing}
Gal, R.; Vinker, Y.; Alaluf, Y.; Bermano, A.~H.; Cohen-Or, D.; Shamir, A.; and Chechik, G. 2023.
\newblock Breathing Life Into Sketches Using Text-to-Video Priors.

\bibitem[{Girdhar et~al.(2024)Girdhar, Singh, Brown, Duval, Azadi, Rambhatla, Shah, Yin, Parikh, and Misra}]{girdhar2024emu}
Girdhar, R.; Singh, M.; Brown, A.; Duval, Q.; Azadi, S.; Rambhatla, S.~S.; Shah, A.; Yin, X.; Parikh, D.; and Misra, I. 2024.
\newblock Emu Video: Factorizing Text-to-Video Generation by Explicit Image Conditioning.
\newblock arXiv:2311.10709.

\bibitem[{Gong et~al.(2023)Gong, Pang, Cun, Xia, He, Chen, Wang, Zhang, Wang, Shan, and Yang}]{gong2023talecrafter}
Gong, Y.; Pang, Y.; Cun, X.; Xia, M.; He, Y.; Chen, H.; Wang, L.; Zhang, Y.; Wang, X.; Shan, Y.; and Yang, Y. 2023.
\newblock TaleCrafter: Interactive Story Visualization with Multiple Characters.
\newblock arXiv:2305.18247.

\bibitem[{G{\"u}ler, Neverova, and Kokkinos(2018)}]{guler2018densepose}
G{\"u}ler, R.~A.; Neverova, N.; and Kokkinos, I. 2018.
\newblock Densepose: Dense human pose estimation in the wild.
\newblock In \emph{Proceedings of the IEEE conference on computer vision and pattern recognition}, 7297--7306.

\bibitem[{Guo et~al.(2023)Guo, Yang, Rao, Agrawala, Lin, and Dai}]{guo2023sparsectrl}
Guo, Y.; Yang, C.; Rao, A.; Agrawala, M.; Lin, D.; and Dai, B. 2023.
\newblock SparseCtrl: Adding Sparse Controls to Text-to-Video Diffusion Models.
\newblock arXiv:2311.16933.

\bibitem[{Gupta et~al.(2023)Gupta, Yu, Sohn, Gu, Hahn, Fei-Fei, Essa, Jiang, and Lezama}]{gupta2023walt}
Gupta, A.; Yu, L.; Sohn, K.; Gu, X.; Hahn, M.; Fei-Fei, L.; Essa, I.; Jiang, L.; and Lezama, J. 2023.
\newblock Photorealistic Video Generation with Diffusion Models.
\newblock arXiv:2312.06662.

\bibitem[{He et~al.(2024)He, Xu, Guo, Wetzstein, Dai, Li, and Yang}]{he2024cameractrl}
He, H.; Xu, Y.; Guo, Y.; Wetzstein, G.; Dai, B.; Li, H.; and Yang, C. 2024.
\newblock CameraCtrl: Enabling Camera Control for Text-to-Video Generation.
\newblock arXiv:2404.02101.

\bibitem[{He et~al.(2022)He, Yang, Zhang, Shan, and Chen}]{he2022lvdm}
He, Y.; Yang, T.; Zhang, Y.; Shan, Y.; and Chen, Q. 2022.
\newblock Latent Video Diffusion Models for High-Fidelity Long Video Generation.

\bibitem[{Ho et~al.(2022{\natexlab{a}})Ho, Chan, Saharia, Whang, Gao, Gritsenko, Kingma, Poole, Norouzi, Fleet, and Salimans}]{ho2022imagenvideohighdefinition}
Ho, J.; Chan, W.; Saharia, C.; Whang, J.; Gao, R.; Gritsenko, A.; Kingma, D.~P.; Poole, B.; Norouzi, M.; Fleet, D.~J.; and Salimans, T. 2022{\natexlab{a}}.
\newblock Imagen Video: High Definition Video Generation with Diffusion Models.
\newblock arXiv:2210.02303.

\bibitem[{Ho, Jain, and Abbeel(2020)}]{ho2020denoising}
Ho, J.; Jain, A.; and Abbeel, P. 2020.
\newblock Denoising diffusion probabilistic models.
\newblock \emph{Advances in neural information processing systems}, 33: 6840--6851.

\bibitem[{Ho and Salimans(2022)}]{ho2022classifier}
Ho, J.; and Salimans, T. 2022.
\newblock Classifier-free diffusion guidance.
\newblock \emph{arXiv preprint arXiv:2207.12598}.

\bibitem[{Ho et~al.(2022{\natexlab{b}})Ho, Salimans, Gritsenko, Chan, Norouzi, and Fleet}]{ho2022video}
Ho, J.; Salimans, T.; Gritsenko, A.; Chan, W.; Norouzi, M.; and Fleet, D.~J. 2022{\natexlab{b}}.
\newblock Video diffusion models.
\newblock \emph{arXiv:2204.03458}.

\bibitem[{Hong et~al.(2022{\natexlab{a}})Hong, Zhang, Pan, Cai, Yang, and Liu}]{hong2022avatarclip}
Hong, F.; Zhang, M.; Pan, L.; Cai, Z.; Yang, L.; and Liu, Z. 2022{\natexlab{a}}.
\newblock AvatarCLIP: Zero-Shot Text-Driven Generation and Animation of 3D Avatars.
\newblock \emph{ACM Transactions on Graphics (TOG)}, 41(4): 1--19.

\bibitem[{Hong et~al.(2022{\natexlab{b}})Hong, Ding, Zheng, Liu, and Tang}]{hong2022cogvideo}
Hong, W.; Ding, M.; Zheng, W.; Liu, X.; and Tang, J. 2022{\natexlab{b}}.
\newblock CogVideo: Large-scale Pretraining for Text-to-Video Generation via Transformers.
\newblock \emph{arXiv preprint arXiv:2205.15868}.

\bibitem[{Hu et~al.(2021)Hu, Shen, Wallis, Allen-Zhu, Li, Wang, Wang, and Chen}]{hu2021lora}
Hu, E.~J.; Shen, Y.; Wallis, P.; Allen-Zhu, Z.; Li, Y.; Wang, S.; Wang, L.; and Chen, W. 2021.
\newblock Lora: Low-rank adaptation of large language models.
\newblock \emph{arXiv preprint arXiv:2106.09685}.

\bibitem[{Hu et~al.(2023)Hu, Gao, Zhang, Sun, Zhang, and Bo}]{hu2023animateanyone}
Hu, L.; Gao, X.; Zhang, P.; Sun, K.; Zhang, B.; and Bo, L. 2023.
\newblock Animate Anyone: Consistent and Controllable Image-to-Video Synthesis for Character Animation.
\newblock \emph{arXiv preprint arXiv:2311.17117}.

\bibitem[{Hu and Xu(2023)}]{hu2023videocontrolnet}
Hu, Z.; and Xu, D. 2023.
\newblock VideoControlNet: A Motion-Guided Video-to-Video Translation Framework by Using Diffusion Model with ControlNet.
\newblock arXiv:2307.14073.

\bibitem[{Huang et~al.(2024)Huang, Shao, Zhang, Zhang, Feng, Liu, and Wang}]{huang2023humannorm}
Huang, X.; Shao, R.; Zhang, Q.; Zhang, H.; Feng, Y.; Liu, Y.; and Wang, Q. 2024.
\newblock Humannorm: Learning normal diffusion model for high-quality and realistic 3d human generation.

\bibitem[{Karaev et~al.(2023)Karaev, Rocco, Graham, Neverova, Vedaldi, and Rupprecht}]{karaev2023cotracker}
Karaev, N.; Rocco, I.; Graham, B.; Neverova, N.; Vedaldi, A.; and Rupprecht, C. 2023.
\newblock Cotracker: It is better to track together.
\newblock \emph{arXiv preprint arXiv:2307.07635}.

\bibitem[{Karras et~al.(2023)Karras, Holynski, Wang, and Kemelmacher-Shlizerman}]{karras2023dreampose}
Karras, J.; Holynski, A.; Wang, T.-C.; and Kemelmacher-Shlizerman, I. 2023.
\newblock DreamPose: Fashion Image-to-Video Synthesis via Stable Diffusion.

\bibitem[{Kerbl et~al.(2023)Kerbl, Kopanas, Leimk{\"u}hler, and Drettakis}]{kerbl20233Dgaussians}
Kerbl, B.; Kopanas, G.; Leimk{\"u}hler, T.; and Drettakis, G. 2023.
\newblock 3D Gaussian Splatting for Real-Time Radiance Field Rendering.
\newblock \emph{ACM Transactions on Graphics}, 42(4).

\bibitem[{Kolotouros et~al.(2023)Kolotouros, Alldieck, Zanfir, Bazavan, Fieraru, and Sminchisescu}]{kolotouros2023dreamhuman}
Kolotouros, N.; Alldieck, T.; Zanfir, A.; Bazavan, E.~G.; Fieraru, M.; and Sminchisescu, C. 2023.
\newblock DreamHuman: Animatable 3D Avatars from Text.

\bibitem[{Lewis, Cordner, and Fong(2023)}]{lewis2023pose}
Lewis, J.~P.; Cordner, M.; and Fong, N. 2023.
\newblock Pose space deformation: a unified approach to shape interpolation and skeleton-driven deformation.
\newblock In \emph{Seminal Graphics Papers: Pushing the Boundaries, Volume 2}, 811--818.

\bibitem[{Li et~al.(2024)Li, Tucker, Snavely, and Holynski}]{li2024generative}
Li, Z.; Tucker, R.; Snavely, N.; and Holynski, A. 2024.
\newblock Generative Image Dynamics.
\newblock In \emph{Proceedings of the IEEE/CVF Conference on Computer Vision and Pattern Recognition (CVPR)}.

\bibitem[{Liang et~al.(2024)Liang, Zhang, Li, Yu, and Xu}]{liang2024intergen}
Liang, H.; Zhang, W.; Li, W.; Yu, J.; and Xu, L. 2024.
\newblock Intergen: Diffusion-based multi-human motion generation under complex interactions.
\newblock \emph{International Journal of Computer Vision}, 1--21.

\bibitem[{Liao et~al.(2024)Liao, Yi, Xiu, Tang, Huang, Thies, and Black}]{liao2024tada}
Liao, T.; Yi, H.; Xiu, Y.; Tang, J.; Huang, Y.; Thies, J.; and Black, M.~J. 2024.
\newblock {TADA! Text to Animatable Digital Avatars}.
\newblock In \emph{International Conference on 3D Vision (3DV)}.

\bibitem[{Liu et~al.(2023{\natexlab{a}})Liu, Li, Wu, and Lee}]{liu2023llava}
Liu, H.; Li, C.; Wu, Q.; and Lee, Y.~J. 2023{\natexlab{a}}.
\newblock Visual Instruction Tuning.

\bibitem[{Liu et~al.(2023{\natexlab{b}})Liu, Zhan, Tang, Shan, Zeng, Lin, Liu, and Liu}]{liu2023humangaussian}
Liu, X.; Zhan, X.; Tang, J.; Shan, Y.; Zeng, G.; Lin, D.; Liu, X.; and Liu, Z. 2023{\natexlab{b}}.
\newblock HumanGaussian: Text-Driven 3D Human Generation with Gaussian Splatting.
\newblock \emph{arXiv preprint arXiv:2311.17061}.

\bibitem[{Long et~al.(2024)Long, Qiu, Yao, and Mei}]{long2024videodrafter}
Long, F.; Qiu, Z.; Yao, T.; and Mei, T. 2024.
\newblock VideoDrafter: Content-Consistent Multi-Scene Video Generation with LLM.
\newblock arXiv:2401.01256.

\bibitem[{Loper et~al.(2015)Loper, Mahmood, Romero, Pons-Moll, and Black}]{SMPL:2015}
Loper, M.; Mahmood, N.; Romero, J.; Pons-Moll, G.; and Black, M.~J. 2015.
\newblock {SMPL}: A Skinned Multi-Person Linear Model.
\newblock \emph{ACM Trans. Graphics (Proc. SIGGRAPH Asia)}, 34(6): 248:1--248:16.

\bibitem[{Ma et~al.(2024)Ma, Wang, Jia, Chen, Liu, Li, Chen, and Qiao}]{ma2024latte}
Ma, X.; Wang, Y.; Jia, G.; Chen, X.; Liu, Z.; Li, Y.-F.; Chen, C.; and Qiao, Y. 2024.
\newblock Latte: Latent Diffusion Transformer for Video Generation.
\newblock \emph{arXiv preprint arXiv:2401.03048}.

\bibitem[{Pavlakos et~al.(2019)Pavlakos, Choutas, Ghorbani, Bolkart, Osman, Tzionas, and Black}]{SMPL-X:2019}
Pavlakos, G.; Choutas, V.; Ghorbani, N.; Bolkart, T.; Osman, A. A.~A.; Tzionas, D.; and Black, M.~J. 2019.
\newblock Expressive Body Capture: {3D} Hands, Face, and Body from a Single Image.
\newblock In \emph{Proceedings IEEE Conf. on Computer Vision and Pattern Recognition (CVPR)}, 10975--10985.

\bibitem[{Peebles and Xie(2022)}]{peebles2022dit}
Peebles, W.; and Xie, S. 2022.
\newblock Scalable Diffusion Models with Transformers.
\newblock \emph{arXiv preprint arXiv:2212.09748}.

\bibitem[{Poole et~al.(2022)Poole, Jain, Barron, and Mildenhall}]{poole2022dreamfusion}
Poole, B.; Jain, A.; Barron, J.~T.; and Mildenhall, B. 2022.
\newblock Dreamfusion: Text-to-3d using 2d diffusion.
\newblock \emph{arXiv preprint arXiv:2209.14988}.

\bibitem[{Radford et~al.(2021)Radford, Kim, Hallacy, Ramesh, Goh, Agarwal, Sastry, Askell, Mishkin, Clark et~al.}]{radford2021learning}
Radford, A.; Kim, J.~W.; Hallacy, C.; Ramesh, A.; Goh, G.; Agarwal, S.; Sastry, G.; Askell, A.; Mishkin, P.; Clark, J.; et~al. 2021.
\newblock Learning transferable visual models from natural language supervision.
\newblock In \emph{International conference on machine learning}, 8748--8763. PMLR.

\bibitem[{Rombach et~al.(2022)Rombach, Blattmann, Lorenz, Esser, and Ommer}]{rombach2022high}
Rombach, R.; Blattmann, A.; Lorenz, D.; Esser, P.; and Ommer, B. 2022.
\newblock High-resolution image synthesis with latent diffusion models.
\newblock In \emph{Proceedings of the IEEE/CVF conference on computer vision and pattern recognition}, 10684--10695.

\bibitem[{Ronneberger, Fischer, and Brox(2015)}]{ronneberger2015u}
Ronneberger, O.; Fischer, P.; and Brox, T. 2015.
\newblock U-net: Convolutional networks for biomedical image segmentation.
\newblock In \emph{Medical image computing and computer-assisted intervention--MICCAI 2015: 18th international conference, Munich, Germany, October 5-9, 2015, proceedings, part III 18}, 234--241. Springer.

\bibitem[{Shan et~al.(2024)Shan, Dong, Han, Yao, Liu, Nwogu, Qi, and Hill}]{shan2024towards}
Shan, M.; Dong, L.; Han, Y.; Yao, Y.; Liu, T.; Nwogu, I.; Qi, G.-J.; and Hill, M. 2024.
\newblock Towards Open Domain Text-Driven Synthesis of Multi-Person Motions.
\newblock \emph{arXiv preprint arXiv:2405.18483}.

\bibitem[{Skorokhodov, Tulyakov, and Elhoseiny(2021)}]{skorokhodov2021styleganv}
Skorokhodov, I.; Tulyakov, S.; and Elhoseiny, M. 2021.
\newblock StyleGAN-V: A Continuous Video Generator with the Price, Image Quality and Perks of StyleGAN2.

\bibitem[{Sohl-Dickstein et~al.(2015)Sohl-Dickstein, Weiss, Maheswaranathan, and Ganguli}]{sohl2015deep}
Sohl-Dickstein, J.; Weiss, E.; Maheswaranathan, N.; and Ganguli, S. 2015.
\newblock Deep unsupervised learning using nonequilibrium thermodynamics.
\newblock In \emph{International conference on machine learning}, 2256--2265. PMLR.

\bibitem[{Song, Meng, and Ermon(2020)}]{song2020denoising}
Song, J.; Meng, C.; and Ermon, S. 2020.
\newblock Denoising diffusion implicit models.
\newblock \emph{arXiv preprint arXiv:2010.02502}.

\bibitem[{Sun et~al.(2023)Sun, Bao, Liu, Mei, and Black}]{sun2023trace}
Sun, Y.; Bao, Q.; Liu, W.; Mei, T.; and Black, M.~J. 2023.
\newblock TRACE: 5D temporal regression of avatars with dynamic cameras in 3D environments.
\newblock In \emph{Proceedings of the IEEE/CVF Conference on Computer Vision and Pattern Recognition}, 8856--8866.

\bibitem[{Tulyakov et~al.(2018)Tulyakov, Liu, Yang, and Kautz}]{tulyakov2018mocogan}
Tulyakov, S.; Liu, M.-Y.; Yang, X.; and Kautz, J. 2018.
\newblock {MoCoGAN}: Decomposing motion and content for video generation.
\newblock In \emph{IEEE Conference on Computer Vision and Pattern Recognition (CVPR)}, 1526--1535.

\bibitem[{Wang et~al.(2023{\natexlab{a}})Wang, Yuan, Chen, Zhang, Wang, and Zhang}]{wang2023modelscope}
Wang, J.; Yuan, H.; Chen, D.; Zhang, Y.; Wang, X.; and Zhang, S. 2023{\natexlab{a}}.
\newblock Modelscope text-to-video technical report.
\newblock \emph{arXiv preprint arXiv:2308.06571}.

\bibitem[{Wang et~al.(2024)Wang, Zhang, Zou, Zeng, Wei, Yuan, and Li}]{wang2024boximator}
Wang, J.; Zhang, Y.; Zou, J.; Zeng, Y.; Wei, G.; Yuan, L.; and Li, H. 2024.
\newblock Boximator: Generating Rich and Controllable Motions for Video Synthesis.
\newblock arXiv:2402.01566.

\bibitem[{Wang et~al.(2021)Wang, Liu, Liu, Theobalt, Komura, and Wang}]{wang2021neus}
Wang, P.; Liu, L.; Liu, Y.; Theobalt, C.; Komura, T.; and Wang, W. 2021.
\newblock NeuS: Learning Neural Implicit Surfaces by Volume Rendering for Multi-view Reconstruction.
\newblock \emph{arXiv preprint arXiv:2106.10689}.

\bibitem[{Wang et~al.(2023{\natexlab{b}})Wang, Li, Lin, Lin, Yang, Zhang, Liu, and Wang}]{wang2023disco}
Wang, T.; Li, L.; Lin, K.; Lin, C.-C.; Yang, Z.; Zhang, H.; Liu, Z.; and Wang, L. 2023{\natexlab{b}}.
\newblock DisCo: Disentangled Control for Referring Human Dance Generation in Real World.
\newblock \emph{arXiv preprint arXiv:2307.00040}.

\bibitem[{Wang et~al.(2022)Wang, Zhang, Zhang, Ouyang, Chen, Chen, and Wen}]{wang2022pretraining}
Wang, T.; Zhang, T.; Zhang, B.; Ouyang, H.; Chen, D.; Chen, Q.; and Wen, F. 2022.
\newblock Pretraining is all you need for image-to-image translation.
\newblock \emph{arXiv preprint arXiv:2205.12952}.

\bibitem[{Xing et~al.(2023)Xing, Xia, Zhang, Chen, Wang, Wong, and Shan}]{xing2023dynamicrafter}
Xing, J.; Xia, M.; Zhang, Y.; Chen, H.; Wang, X.; Wong, T.-T.; and Shan, Y. 2023.
\newblock DynamiCrafter: Animating Open-domain Images with Video Diffusion Priors.

\bibitem[{Xu et~al.(2024{\natexlab{a}})Xu, Nie, Liu, Liu, Kautz, Wang, and Vahdat}]{xu2024camco}
Xu, D.; Nie, W.; Liu, C.; Liu, S.; Kautz, J.; Wang, Z.; and Vahdat, A. 2024{\natexlab{a}}.
\newblock CamCo: Camera-Controllable 3D-Consistent Image-to-Video Generation.
\newblock \emph{arXiv preprint arXiv:2406.02509}.

\bibitem[{Xu et~al.(2024{\natexlab{b}})Xu, Zhang, Liew, Yan, Liu, Zhang, Feng, and Shou}]{xu2023magicanimate}
Xu, Z.; Zhang, J.; Liew, J.~H.; Yan, H.; Liu, J.-W.; Zhang, C.; Feng, J.; and Shou, M.~Z. 2024{\natexlab{b}}.
\newblock MagicAnimate: Temporally Consistent Human Image Animation using Diffusion Model.

\bibitem[{Yang et~al.(2024{\natexlab{a}})Yang, Hou, Huang, Ma, Wan, Zhang, Chen, and Liao}]{yang2024direct}
Yang, S.; Hou, L.; Huang, H.; Ma, C.; Wan, P.; Zhang, D.; Chen, X.; and Liao, J. 2024{\natexlab{a}}.
\newblock Direct-a-Video: Customized Video Generation with User-Directed Camera Movement and Object Motion.
\newblock \emph{arXiv preprint arXiv:2402.03162}.

\bibitem[{Yang et~al.(2023)Yang, Zhou, Liu, , and Loy}]{yang2023rerender}
Yang, S.; Zhou, Y.; Liu, Z.; ; and Loy, C.~C. 2023.
\newblock Rerender A Video: Zero-Shot Text-Guided Video-to-Video Translation.
\newblock In \emph{ACM SIGGRAPH Asia 2023 Conference Proceedings}.

\bibitem[{Yang et~al.(2024{\natexlab{b}})Yang, Teng, Zheng, Ding, Huang, Xu, Yang, Zhang, Gu, Feng, Yin, Hong, Wang, Cheng, Zhang, Liu, Xu, Dong, and Tang}]{yang2024cogvideox}
Yang, Z.; Teng, J.; Zheng, W.; Ding, M.; Huang, S.; Xu, J.; Yang, Y.; Zhang, X.; Gu, X.; Feng, G.; Yin, D.; Hong, W.; Wang, W.; Cheng, Y.; Zhang, Y.; Liu, T.; Xu, B.; Dong, Y.; and Tang, J. 2024{\natexlab{b}}.
\newblock CogVideoX: Text-to-Video Diffusion Models with An Expert Transformer.

\bibitem[{Yatim et~al.(2024)Yatim, Fridman, Bar-Tal, Kasten, and Dekel}]{yatim2024space}
Yatim, D.; Fridman, R.; Bar-Tal, O.; Kasten, Y.; and Dekel, T. 2024.
\newblock Space-time diffusion features for zero-shot text-driven motion transfer.
\newblock In \emph{Proceedings of the IEEE/CVF Conference on Computer Vision and Pattern Recognition}, 8466--8476.

\bibitem[{Yin et~al.(2023)Yin, Wu, Liang, Shi, Li, Ming, and Duan}]{yin2023dragnuwa}
Yin, S.; Wu, C.; Liang, J.; Shi, J.; Li, H.; Ming, G.; and Duan, N. 2023.
\newblock DragNUWA: Fine-grained Control in Video Generation by Integrating Text, Image, and Trajectory.
\newblock arXiv:2308.08089.

\bibitem[{Yu et~al.(2023)Yu, Cheng, Sohn, Lezama, Zhang, Chang, Hauptmann, Yang, Hao, Essa, and Jiang}]{yu2023magvit}
Yu, L.; Cheng, Y.; Sohn, K.; Lezama, J.; Zhang, H.; Chang, H.; Hauptmann, A.~G.; Yang, M.-H.; Hao, Y.; Essa, I.; and Jiang, L. 2023.
\newblock {MAGVIT}: Masked generative video transformer.
\newblock In \emph{Proceedings of the IEEE/CVF Conference on Computer Vision and Pattern Recognition}.

\bibitem[{Yu et~al.(2022)Yu, Tack, Mo, Kim, Kim, Ha, and Shin}]{yu2022digan}
Yu, S.; Tack, J.; Mo, S.; Kim, H.; Kim, J.; Ha, J.-W.; and Shin, J. 2022.
\newblock Generating Videos with Dynamics-aware Implicit Generative Adversarial Networks.
\newblock In \emph{International Conference on Learning Representations}.

\bibitem[{Zhang et~al.(2023)Zhang, Chen, Yang, Qu, Wang, Chen, Long, Zhu, Du, and Zheng}]{zhang2023avatarverse}
Zhang, H.; Chen, B.; Yang, H.; Qu, L.; Wang, X.; Chen, L.; Long, C.; Zhu, F.; Du, K.; and Zheng, M. 2023.
\newblock AvatarVerse: High-quality and Stable 3D Avatar Creation from Text and Pose.
\newblock arXiv:2308.03610.

\bibitem[{Zhu et~al.(2024)Zhu, Chen, Dai, Xu, Cao, Yao, Zhu, and Zhu}]{zhu2024champ}
Zhu, S.; Chen, J.~L.; Dai, Z.; Xu, Y.; Cao, X.; Yao, Y.; Zhu, H.; and Zhu, S. 2024.
\newblock Champ: Controllable and Consistent Human Image Animation with 3D Parametric Guidance.
\newblock arXiv:2403.14781.

\end{thebibliography}

\appendix

\begin{figure} [t]
\includegraphics[width=\linewidth]{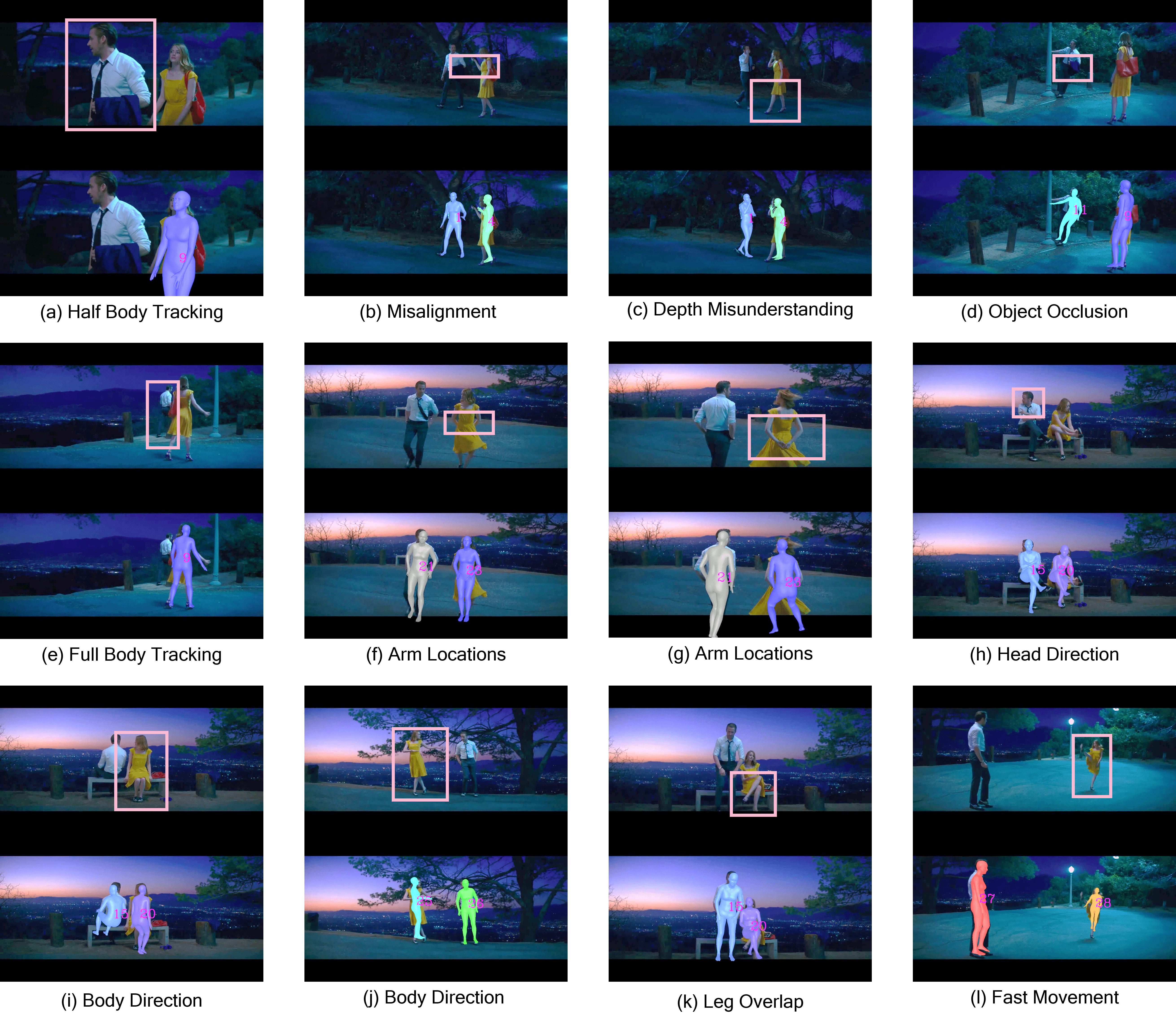}
\centering
\caption{
\textbf{Common Reconstruction Errors in TRACE.}
From panel (a) to (l), we display some common errors observed in the output of TRACE~\cite{sun2023trace}. 
The upper row in each panel shows the input frame, while the lower row presents the output rendered by projecting the SMPL~\cite{SMPL:2015} mesh back onto the image plane.
The error regions are highlighted with pink boxes in the upper row. 
The corresponding regions in the lower row are not marked for clarity. 
The numbers on the mesh represent identity numbers generated by TRACE.
}
\label{fig:trace}
\end{figure}

\section{More Data Processing Details}

We selected a well-known clip from the film \textit{La La Land}. This clip has a duration of 4 minutes and 42 seconds. The URL for the clip is provided here\footnote{\url{https://www.youtube.com/watch?v=LM0_hstiMLw&t=5s}}. The downloaded resolution of the clip is 1280 × 720.

\subsection{Preliminaries}

\noindent\textbf{SMPL-X}~\cite{SMPL-X:2019} is a comprehensive 3D parametric human model that defines the shape topology of the body, hands, and face. 
The model consists of 10,475 vertices and 54 keypoints. 
By leveraging pose parameters $\theta$ (which includes body pose $\theta_b$, jaw pose $\theta_f$, and finger pose $\theta_h$), shape parameters $\beta$, and expression parameters $\psi$, the 3D SMPL-X human model $M(\beta, \theta, \psi)$ can be expressed as:
\begin{equation}
T(\beta, \theta, \psi) = \bar{T} + B_s(\beta) + B_p(\theta) + B_e(\psi),
\end{equation}
\begin{equation}
M(\beta, \theta, \psi) = \text{LBS}(T(\beta, \theta, \psi), J(\beta), \theta, \mathcal{W}),
\end{equation}
where $\bar{T}$ denotes the mean template shape; $B_s$, $B_p$, and $B_e$ represent the blend shape functions corresponding to shape, pose, and expression, respectively; $T(\beta, \theta, \psi)$ is the non-rigid deformation of $\bar{T}$; and $\text{LBS}(\cdot)$ is the linear blend skinning function~\cite{lewis2023pose} that maps $T(\beta, \theta, \psi)$ to the target pose $\theta$, using the skeleton joints $J(\beta)$ and the blend weights $\mathcal{W}$ associated with each vertex.

\subsection{TRACE Results}
\begin{figure} [t]
\includegraphics[width=\linewidth]{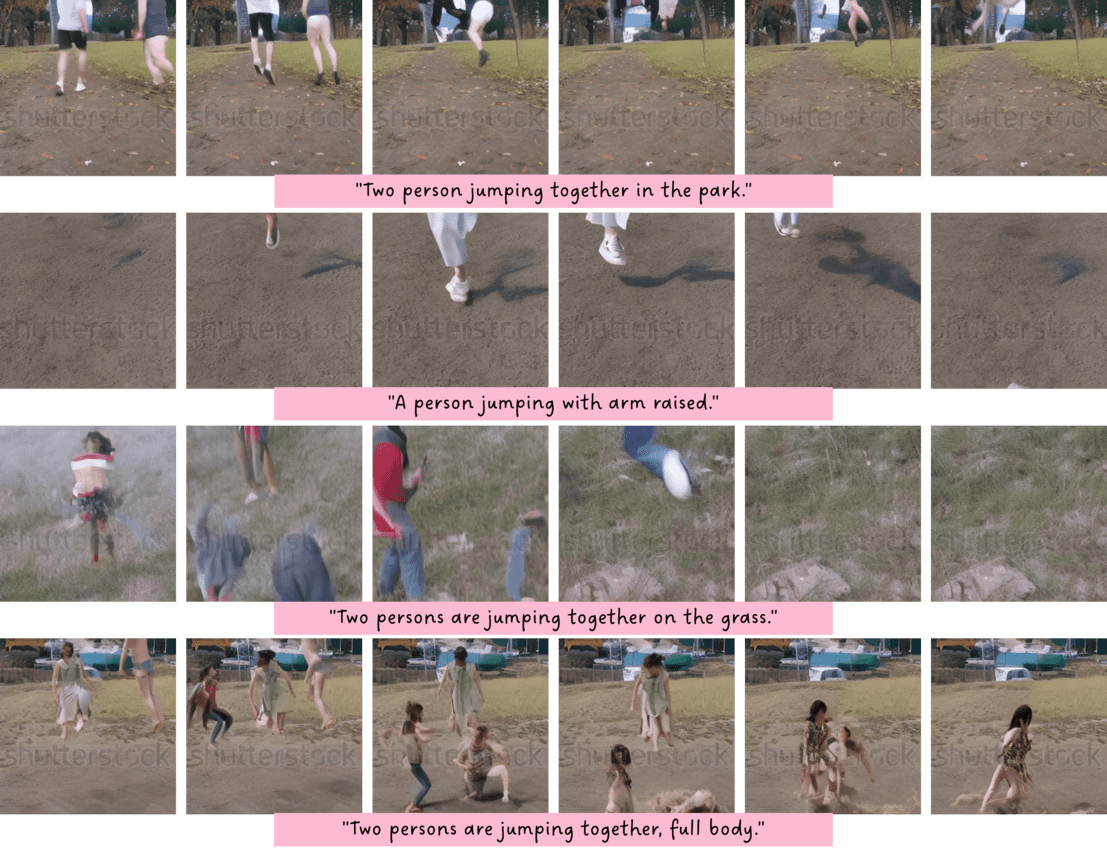}
\centering
\caption{
\textbf{Jumping Results Sampled from ModelScope.}
We present four samples generated by ModelScope~\cite{wang2023modelscope}. 
The corresponding prompt used in ModelScope is shown in the pink box. 
Each sampled video has a resolution of 256 by 256 pixels and a length of 16 frames. We display six evenly distributed frames from the sampled videos.
}
\label{fig:jump}
\end{figure}
\begin{figure} [t]
\includegraphics[width=\linewidth]{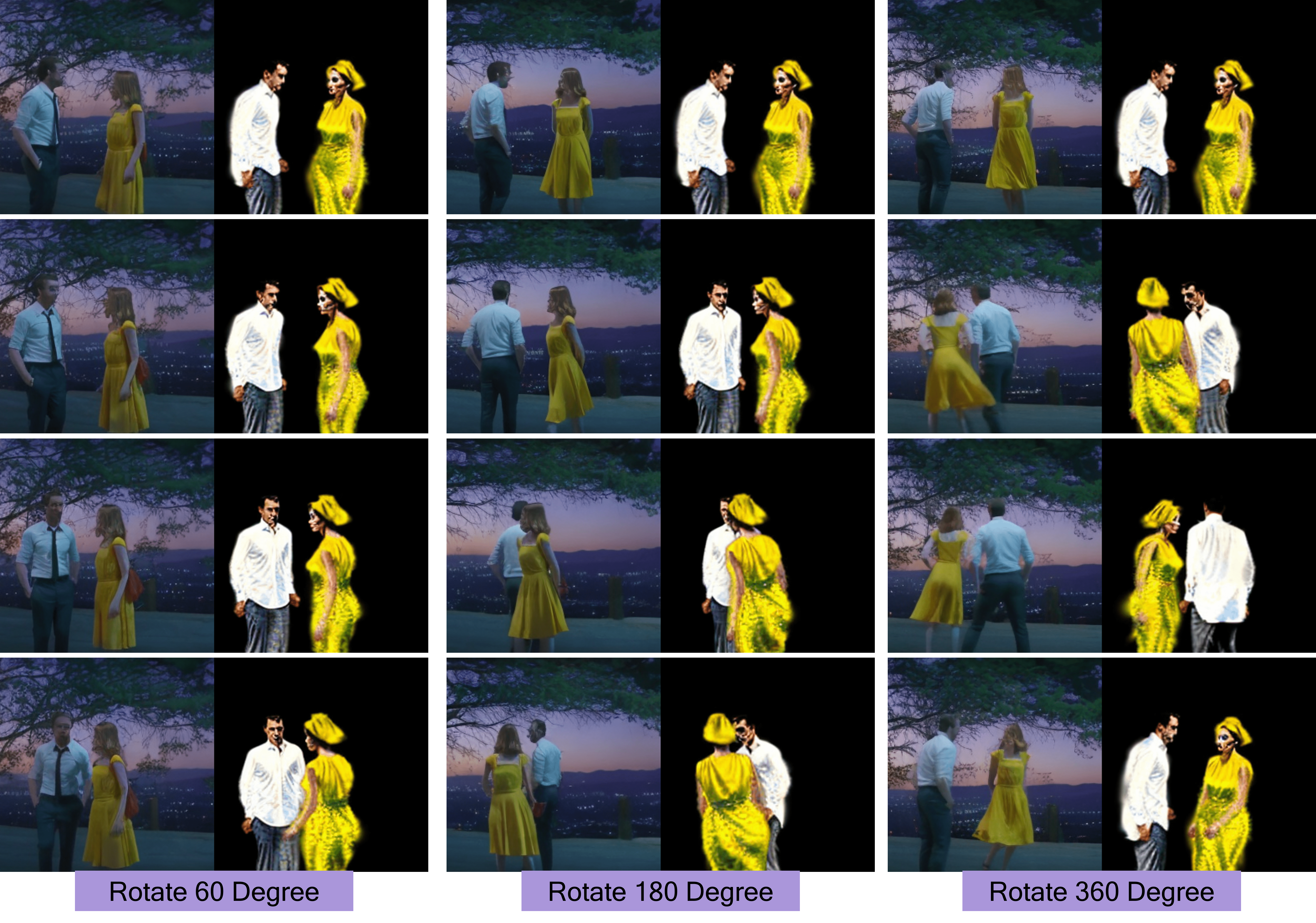}
\centering
\caption{
\textbf{Different Total Camera Rotation Angle in Generated Video.}
We display three different total camera rotation angles: 60, 180, and 360 degrees. By "total angle," we refer to the degree of rotation between the first frame and the last frame.
}
\label{fig:pan}
\end{figure}

\begin{figure*} [t]
\includegraphics[width=\linewidth]{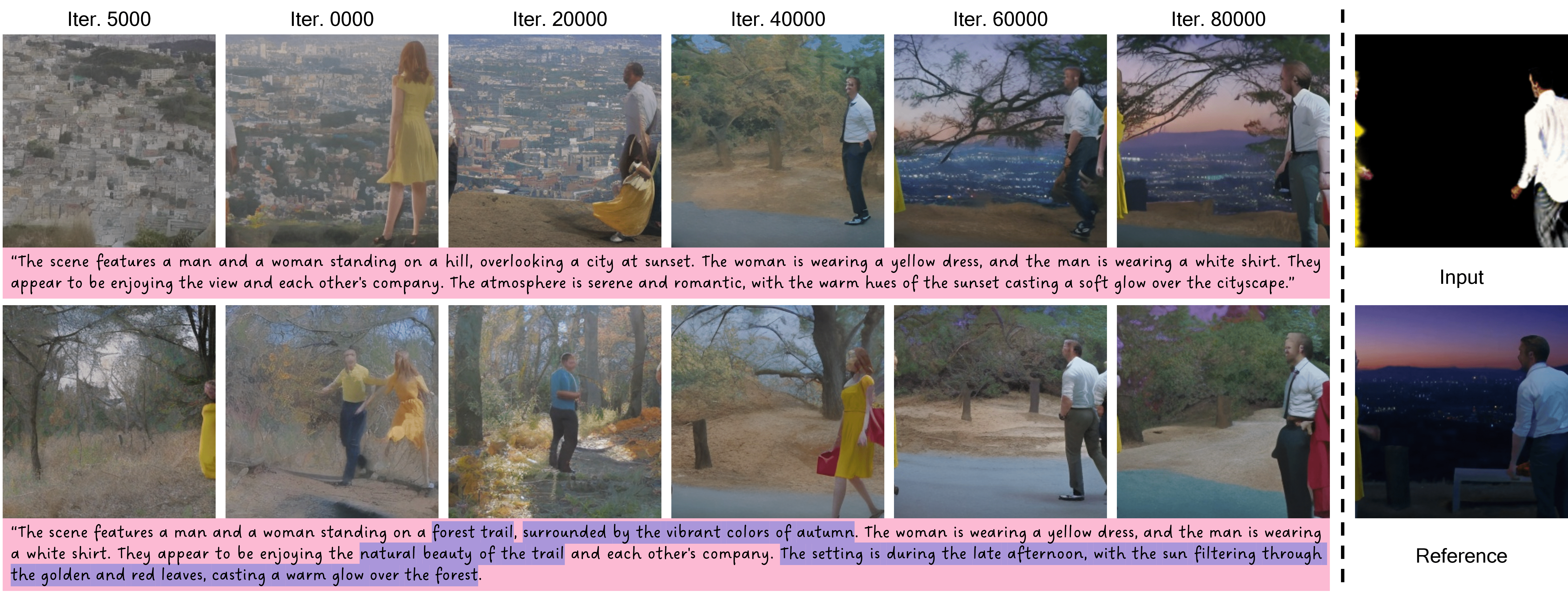}
\centering
\caption{
The visualization illustrates the evolution of the model's behavior as training iterations increase. The left section displays the evaluation of the model's performance: the top row presents the results on the hold-out test data, while the bottom row shows the results when the background prompt is altered. The right section contains the input avatar video frame and the corresponding video frame from the dataset.
}
\label{fig:epoch}
\end{figure*}

Even though TRACE~\cite{sun2023trace} is renowned for robust human trajectory reconstruction, we observed several issues in its application, particularly when used in real-world scenarios. In Fig.~\ref{fig:trace}, we highlight some common mistakes encountered with TRACE in the wild.

As shown in panels (a) and (e), the tracking often fails under certain conditions. 
The first issue arises from TRACE's insufficient handling of partial body tracking. 
The second issue occurs frequently in the presence of occlusions.
In panel (b), the reconstructed right arm appears well-posed but misaligned with the actual position of the actress when projected back onto the image. 
In panel (c), the model struggles with depth perception, making it difficult to accurately distinguish the front and back positions of the left and right legs. 
In panel (d), the man is carrying a blazer on his left forearm, but the algorithm fails to recognize this, leading to incorrect localization of the forearm.

Panels (f) and (g) show instances where the actress’s arms are positioned behind her back, yet the algorithm incorrectly places the arms in front. In panel (h), a noticeable head-twisting action is inaccurately represented, likely due to the limitations of the SMPL model, which constrains neck rotation. 
Panels (i) and (j) illustrate errors in body orientation, potentially due to the algorithm's disregard for temporal information. In panel (k), the algorithm fails to accurately locate overlapping legs, possibly due to limitations in the SMPL model. 
Finally, in panel (l), a quick high-kick action by the actress is not properly recognized by the algorithm. Although such actions occur at a lower frequency, they cannot be overlooked, especially given the importance of action films in the film industry.

We hope these observations will stimulate further advancements in human trajectory reconstruction within our community. A more detailed analysis of how these issues affect the generated results are provided in later sections.

\subsection{Prompt Used in HumanGaussian}

The input prompts used in HumanGaussian are as follows:

\texttt{A woman wearing a bright yellow dress, which stands out against the darker background. She is walking with a relaxed posture, and her dress is sleeveless, flowing just above her knees.}

\texttt{A man wearing a white dress shirt and dark pants, suggesting a formal or semi-formal appearance.}

\subsection{Converting TRACE to HumanGaussian}

The output from TRACE~\cite{sun2023trace} is provided in SMPL~\cite{SMPL:2015} format, whereas the A-pose template in HumanGaussian~\cite{liu2023humangaussian} utilizes the SMPL-X~\cite{SMPL-X:2019} format. 
To align these formats, we convert the SMPL A-pose to the SMPL-X A-pose using the conversion tools available in the SMPL-X repository.\footnote{\url{https://github.com/vchoutas/smplx/blob/main/transfer_model/README.md}}
Once trained, HumanGaussian can be driven by skeletal motion from any source that shares the same manifold, even if the topologies differ. Consequently, this allows the output from TRACE to effectively drive the HumanGaussian model.

\section{More Results}

\subsection{More Qualitative Analysis}
\label{sec:jump}

It is worth noting that in some cases, such as with the jumping motion, the avatar raises its hands while jumping, but this hand-raising action is not accurately replicated in the generated video.
There are two possible explanations for this discrepancy. 
First, most of the jumping actions in the training data may have been captured without the hand-raising component, leading the model to associate jumping primarily with non-hand-raising motions. 
Second, the model may not focus as effectively on arm details, as arms are typically thinner and less prominent compared to the body and legs. 

To further investigate the first reason, we sampled additional jumping videos from ModelScope~\cite{wang2023modelscope}, as shown in Fig.~\ref{fig:jump}. 
These results indicate that the generated videos predominantly focus on the lower body during the jumping motion. 
Even when we included prompt phrases like \texttt{"full body"} in the input, the resulting videos still failed to depict the avatar with raised arms during the jump. 
Moreover, when we explicitly prompted the model with \texttt{"jumping with arms raised"}, the generated videos continued to emphasize the lower body, particularly the legs and feet. 
This suggests that the model struggles to accurately interpret and generate specific details related to certain actions, such as \texttt{"jumping with arms raised"}, indicating a limitation in its understanding of such concepts.
Additionally, the model also struggles with accurately controlling the number of people in the generated video, an issue that our proposed method effectively addresses.

\subsection{Camera Movement}
\begin{figure} [t]
\includegraphics[width=\linewidth]{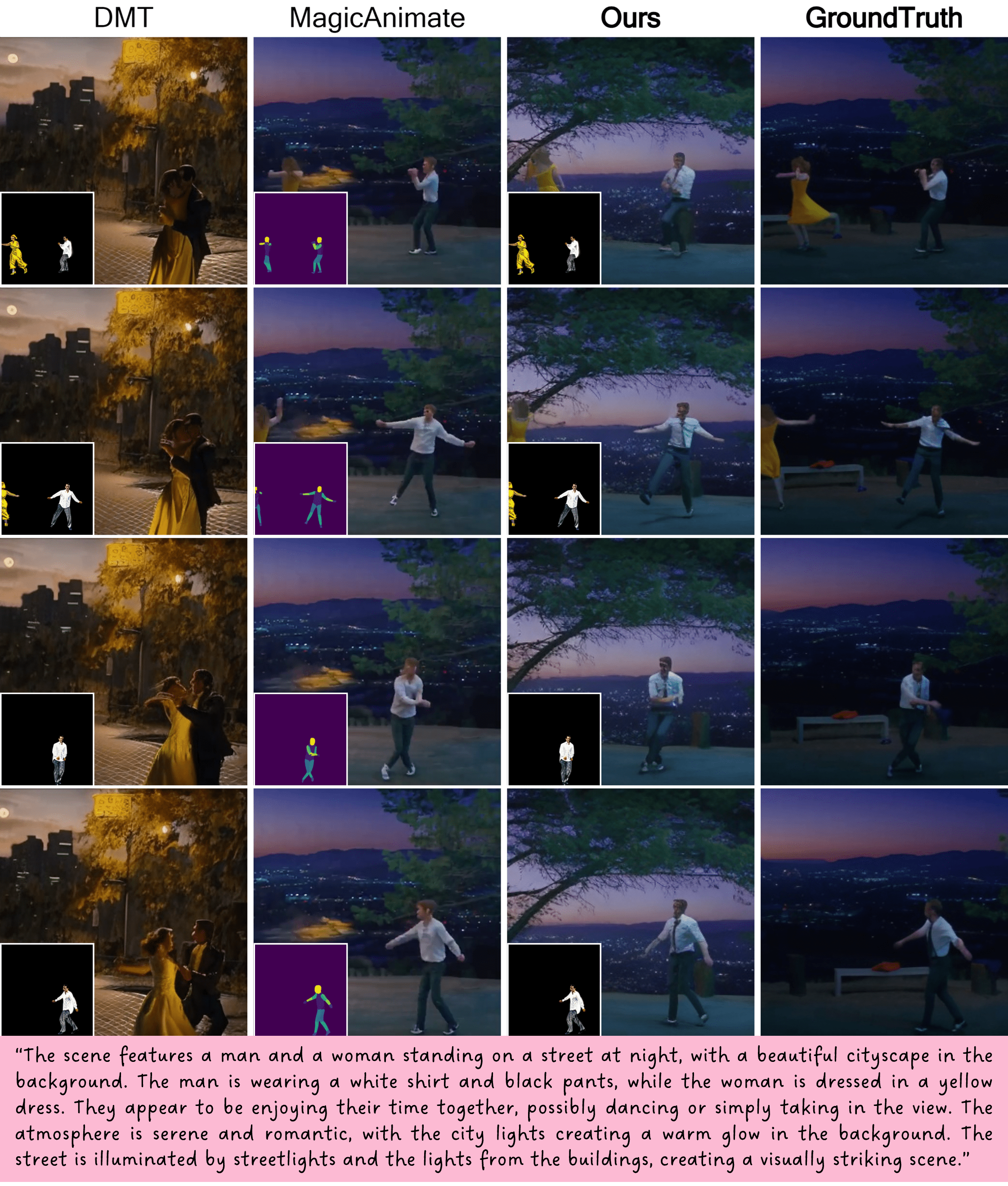}
\centering
\caption{
\textbf{More Baseline Comparison Results.}
}
\label{fig:baseline_11}
\end{figure}
\begin{figure} [t]
\includegraphics[width=\linewidth]{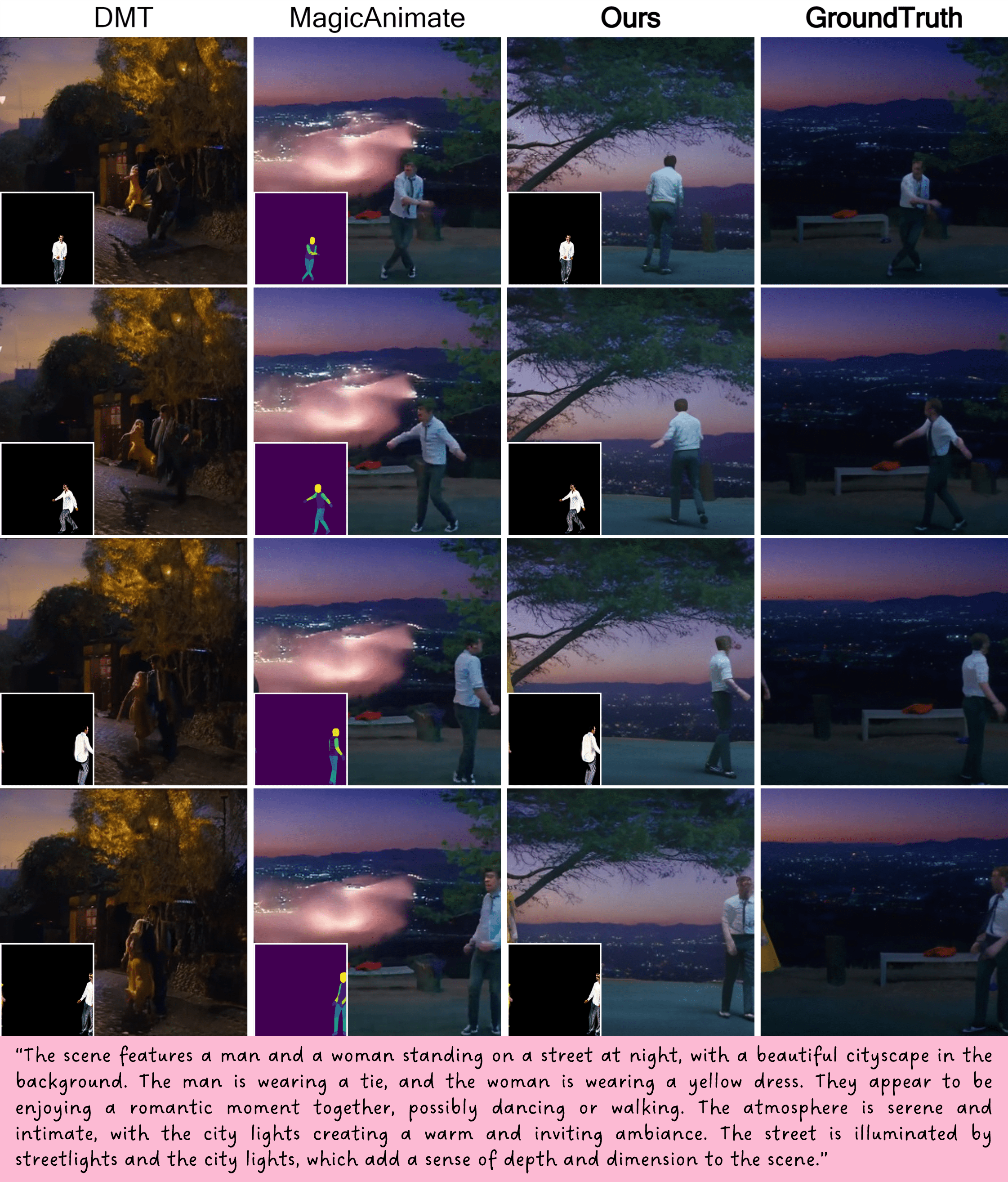}
\centering
\caption{
\textbf{More Baseline Comparison Results.}
}
\label{fig:baseline_21}
\end{figure}
\begin{figure} [t]
\includegraphics[width=\linewidth]{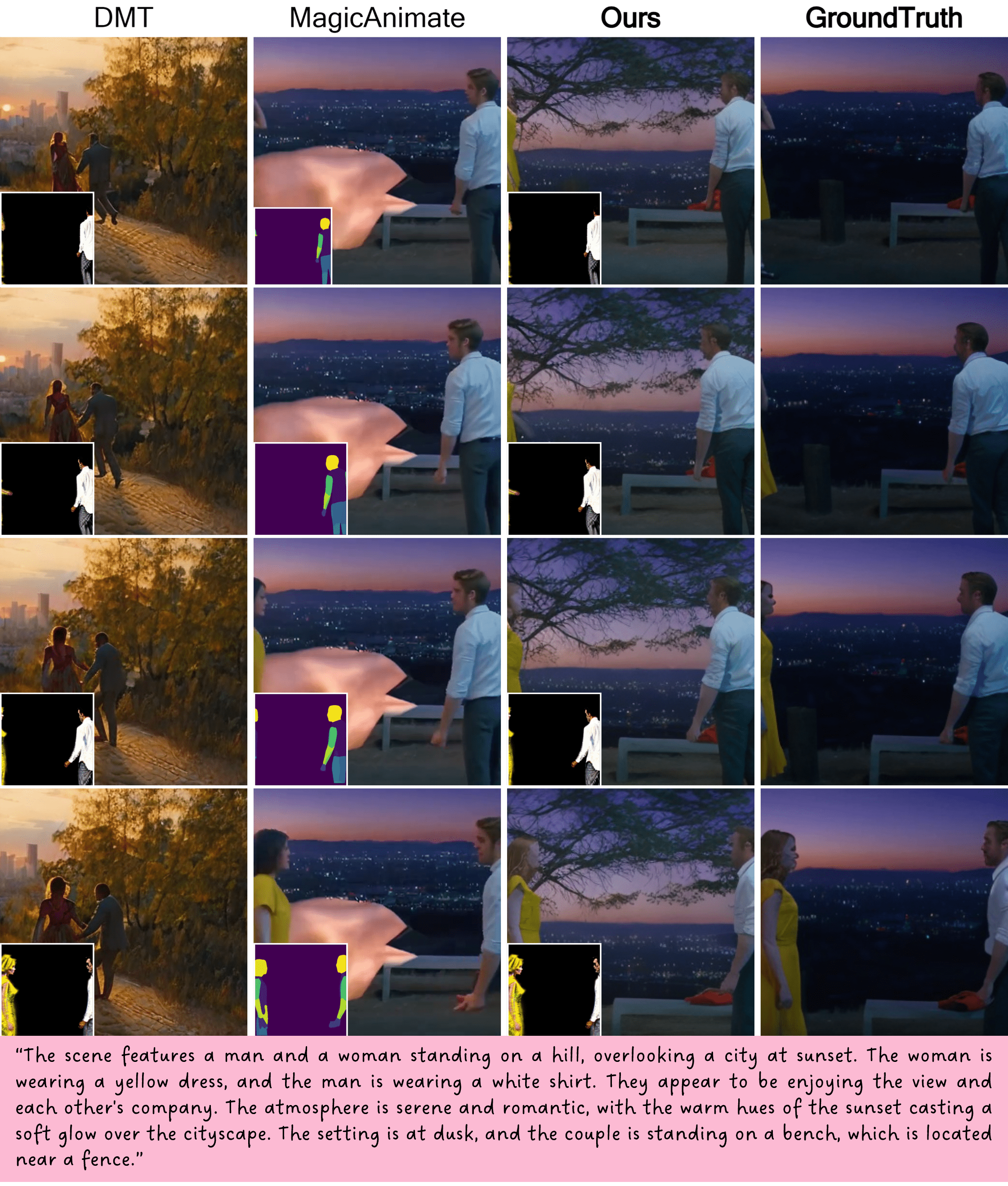}
\centering
\caption{
\textbf{More Baseline Comparison Results.}
}
\label{fig:baseline_31}
\end{figure}

In Fig.~\ref{fig:pan}, we demonstrate the effects of increased camera movement. It is important to note that while the movement in the generated video corresponds to the input condition, the outcome is not fully as desired. The avatar video is produced by rotating the camera around the avatar; however, the generated video interprets camera motion as human body motion while keeping the background static.

This discrepancy suggests a potential future direction: enabling the network to comprehend and accurately represent camera movement in the output. One possible solution is to expose the network to the motion of the world floor, allowing it to implicitly learn camera movement. This would enable the input condition to fully control the camera's motion in the generated video.

\subsection{Baseline}

In Fig.~\ref{fig:baseline_11}, Fig.~\ref{fig:baseline_21}, and Fig.~\ref{fig:baseline_31}, we present additional baseline results for the remaining three test cases.
The results indicate that our method consistently produces higher-quality generated videos. In contrast, MagicAnimate~\cite{xu2023magicanimate} exhibits issues, particularly in generating red regions. We believe this problem arises due to the input condition, where the region is too dark, leading to numerical issues. Since MagicAnimate is primarily trained on datasets with light backgrounds, its performance deteriorates in darker regions.

\subsection{Iterative Performance and Generalization}

In Fig.~\ref{fig:epoch}, we present the model's performance on the test data across different training iterations. 
To evaluate the generalizability of the model, we also modify the background prompt to observe its performance under varying conditions. In the early stages of training, the model struggles to accurately follow the input motion condition. However, as training progresses, the model improves in aligning with the input motion. Conversely, as the number of iterations increases, the model faces the risk of gradually forgetting previously learned knowledge, which may hinder its generalization ability. In our experiments, we selected the model at iteration 60,000 for evaluation, as it represents a trade-off between accurate motion following and generalization.

\subsection{Metrics}

\noindent\underline{CLIP text similarity} computes 
$$
100 * \frac{\mathbf{t} \cdot \mathbf{i}}{\|\mathbf{t}\| \|\mathbf{i}\|}
$$

where
\( \mathbf{t} \) is the embedding vector for the text,
\( \mathbf{i} \) is the embedding vector for the image,
\( \mathbf{t} \cdot \mathbf{i} \) is the dot product of the two vectors,
\( \|\mathbf{t}\| \) and \( \|\mathbf{i}\| \) are the magnitudes (or Euclidean norms) of the text and image embedding vectors, respectively.

\noindent\underline{Motion Fidelity Score} is introduced in ~\cite{yatim2024space} as a metric to evaluate the similarity between tracklets in input and output videos using an off-the-shelf tracking method~\cite{karaev2023cotracker}. 
To estimate the sets of tracklets $\mathcal{T} = \{\tau_1, \ldots, \tau_n\}$ from the input video and $\tilde{\mathcal{T}} = \{\tilde{\tau}_1, \ldots, \tilde{\tau}_m\}$ from the output video. Motion Fidelity Score is inspired by the Chamfer distance. This score is calculated by measuring the similarity between each tracklet in $\mathcal{T}$ and its nearest neighbor in $\tilde{\mathcal{T}}$, and vice versa:

\[
\frac{1}{m} \sum_{\tilde{\tau} \in \tilde{\mathcal{T}}} \max_{\tau \in \mathcal{T}} \text{corr}(\tau, \tilde{\tau}) + \frac{1}{n} \sum_{\tau \in \mathcal{T}} \max_{\tilde{\tau} \in \tilde{\mathcal{T}}} \text{corr}(\tau, \tilde{\tau})
\tag{5}
\]

The correlation between two tracklets, $\text{corr}(\tau, \tilde{\tau})$, is computed as:

\[
\text{corr}(\tau, \tilde{\tau}) = \frac{1}{F} \sum_{k=1}^{F} \frac{v_k^x \cdot \tilde{v}_k^x + v_k^y \cdot \tilde{v}_k^y}{\sqrt{(v_k^x)^2 + (v_k^y)^2} \cdot \sqrt{(\tilde{v}_k^x)^2 + (\tilde{v}_k^y)^2}}
\]

where $(v_k^x, v_k^y)$ and $(\tilde{v}_k^x, \tilde{v}_k^y)$ represent the $k$th frame displacements of tracklets $\tau$ and $\tilde{\tau}$, respectively.

\end{document}